\documentclass[11pt, letterpaper]{article}
\usepackage[utf8]{inputenc}

\usepackage[margin=1in]{geometry}
\usepackage{setspace} 
\usepackage{apacite}
\usepackage{parskip}
\usepackage{natbib}
\usepackage{url} 
\usepackage[hidelinks]{hyperref}
\usepackage{amsmath,amsfonts,amssymb,amsthm}
\usepackage{commath}
\usepackage{graphicx}

\usepackage{bbm} 

\usepackage{enumitem} 

\usepackage{graphicx}      
\usepackage{subcaption}    
\usepackage{lipsum}

\usepackage{algorithm} 
\usepackage{algpseudocode}

\usepackage{verbatim}
\usepackage{listings}

\usepackage{booktabs} 
\usepackage{siunitx} 
\usepackage{array} 
\usepackage{makecell} 
\usepackage{tabularx} 

\usepackage{cleveref} 
\crefname{figure}{}{} 

\usepackage{xr}
\usepackage[affil-it]{authblk}

\numberwithin{equation}{section} 

\title{\vspace{-3em} \textbf{\Large Why You Should Not Trust Interpretations in Machine Learning: Adversarial Attacks on Partial Dependence Plots} \vspace{-0.75em}}

\author[1]{Xi Xin}
\author[2]{Giles Hooker}
\author[1]{Fei Huang\footnote{Correspondence: Xi Xin, xi.xin@unsw.edu.au \\  Giles Hooker, ghooker@wharton.upenn.edu \\   Fei Huang, feihuang@unsw.edu.au.}}

\affil[1]{UNSW Sydney,
School of Risk and Actuarial Studies}
\affil[2]{University of Pennsylvania, Wharton School
Department of Statistics and Data Science}

\date{}

\begin{document}
\onehalfspacing

\maketitle

\vspace{-3em}

\begin{abstract}


The adoption of artificial intelligence (AI) across industries has led to the widespread use of complex black-box models and interpretation tools for decision making. This paper proposes an adversarial framework to uncover the vulnerability of permutation-based interpretation methods for machine learning tasks, with a particular focus on partial dependence (PD) plots. This adversarial framework modifies the original black box model to manipulate its predictions for instances in the extrapolation domain. As a result, it produces deceptive PD plots that can conceal discriminatory behaviors while preserving most of the original model's predictions. This framework can produce multiple fooled PD plots via a single model. By using real-world datasets including an auto insurance claims dataset and COMPAS (Correctional Offender Management Profiling for Alternative Sanctions)  dataset, our results show that it is possible to intentionally hide the discriminatory behavior of a predictor and make the black-box model appear neutral through interpretation tools like PD plots while retaining almost all the predictions of the original black-box model. Managerial insights for regulators and practitioners are provided based on the findings.
\end{abstract}


\section{Introduction}
In recent years, industries have increasingly adopted artificial intelligence (AI), leading to the widespread use of complex AI models. These AI applications can enhance efficiency and accuracy, resulting in time and cost savings. However, these models often operate as ``black boxes'', meaning while we can observe their inputs and outputs, their inner workings remain opaque. This lack of transparency has raised concerns from both regulators and consumers, especially when black-box AI models are used in critical decision-making scenarios.

In this paper, we propose an adversarial framework to demonstrate the susceptibility of partial dependence (PD) plots to adversarial attacks, specifically revealing how these plots can be manipulated by exploiting the extrapolation behavior of correlated features and the aggregation of heterogeneous effects during the averaging process. This adversarial framework modifies the original black box model to manipulate its predictions for instances in the extrapolation domain. As a result, it produces deceptive PD plots that can conceal discriminatory behaviors while preserving most of the original model's predictions. Empirical insurance and COMPAS (Correctional Offender Management Profiling for Alternative Sanctions)  datasets are applied to demonstrate the effectiveness of this model. Our results show that it is possible to intentionally hide the discriminatory behavior of a predictor and make the black-box model appear neutral through interpretation tools like PD plots while retaining almost all the predictions of the original black-box model.

To gain insights into the relationships between model inputs and outputs, various interpretation methods from the growing field of interpretable machine learning can be potentially employed to support the application of black-box models. For example, some of these interpretation methods have garnered attention within sectors such as insurance \citep{kuo2020towards, SOA2021interpretable, delcaillau2022model}, credit scoring \citep{bucker2022transparency,szepannek2023much} or healthcare \citep{mohanty2022comprehensive} in recent years. These methods can be broadly categorized into three groups \citep{SOA2021interpretable}: 1) feature importance, 2) methods for understanding the relationships between model inputs and outputs (main effects), and 3) methods for identifying and visualizing interaction effects. PD plots are suggested as interpretation tools by regulators \citep{NAIC2020regulatory} and practitioners \citep{SOA2021interpretable} in the insurance industry. They are frequently employed in the literature when black-box insurance models are implemented or proposed -- ranging from insurance \citep{guelman2012gradient, yang2018insurance, lee2018delta, xie2021improving, henckaerts2021boosting} to customer churn \citep{lemmens2006bagging, matuszelanski2022customer}, and criminal justice \citep{berk2013statistical}.

The interpretability, explainability, and transparency of black-box algorithms are commonly discussed ethical challenges across various AI principle documents \citep[p.149]{HAI2019indexreport}. Nevertheless, the definitions and interpretations of these principles tend to vary across scientific disciplines and among different stakeholders, including policymakers, technical standardization communities, legal scholars, and applied AI practitioners \citep{panigutti2023role}. Specifically, the EU's General Data Protection Regulation (GDPR) has been interpreted by some as containing a “right to explanation,” a topic which continues to spark debate regarding the legal existence and the feasibility of such a right under the GDPR \citep{wachter2017right, bordt2022post}.


A recent regulatory development in the insurance domain is the expansion of the National Association of Insurance Commissioners' Predictive Models White Paper \citep{NAIC2022trees} in the US insurance industry. It has extended its scope from primarily reviewing rate filings based on Generalized Linear Models (GLMs) \citep{NAIC2020regulatory} to include tree-based models like random forests and gradient boosting machines. The White Paper provides state insurance regulators with comprehensive guidelines for assessing predictive models. Under these guidelines, insurers using tree-based methods in rate filings are expected to implement interpretability plots to describe the relationship between each predictor variable and the target variable, which could range from frequency and severity to loss costs and expenses. Insurers should offer rational explanations for the observed relationship, and are also recommended to obtain variable importance plots to identify significant variables influencing the model's outcomes.

Although stakeholders may possess a certain level of understanding regarding the limitations of these interpretation methods, there is often a lack of awareness regarding the inherent vulnerabilities of these methods. A growing body of literature emphasizes the need for caution in the use of these methods, as they can be unreliable and prone to provide misleading information. For instance, \citet{rudin2019stop} asserts that these explanations are fundamentally flawed as they can never be fully faithful to the original model. \citet{hooker2021unrestricted} demonstrate the vulnerability of permutation-based interpretation methods due to the extrapolation behavior of correlated features. \citet{molnar2022general} uncover and review the general pitfalls of model-agnostic interpretation methods.

Recent years have seen studies developing adversarial attacks of certain interpretation methods, such as LIME, SHAP \citep{slack2020fooling, baniecki2022manipulating, laberge2022fooling} and partial dependence plots \citep{baniecki2023fooling}. Building upon this line of research, we present the vulnerability of permutation-based interpretation methods via adversarial attacks on partial dependence (PD) plots -- a widely utilized interpretation method. There is limited existing literature on developing an adversarial framework to fool PD plots. The only paper we could find is a recent study by \citet{baniecki2023fooling}. Compared to their model which relies on a poisoned dataset to manipulate the PD plots, our framework modifies the model instead, which can achieve substantial fooling effectiveness and allow auditors (regulators) to examine the datasets used.  Our model also extends \citet{slack2020fooling}'s scaffolded classifier to a scaffolded regressor framework. Notably, \citet{slack2020fooling} directly used the permuted data generated by LIME or kernel SHAP in their adversarial framework, however, our approach uses only extrapolated permuted PD data as compensating outputs. This subset, which falls within the extrapolation domain, is less likely to be observed in real data and represents only a portion of all permuted data generated for PD plots. This specificity enhances the efficiency and accuracy of the fooling process, as opposed to using the entire set of permuted data.

Our findings raise concerns about the use of permutation-based interpretation methods for machine learning tasks that require interpretation. This is because the discriminatory behavior of a predictor can be intentionally hidden by tools such as PD plots which make the model appear neutral while preserving nearly all original discriminatory predictions.

The rest of the paper is organized as follows. Section 2 reviews related studies. Section 3 introduces PD plots, discussing both their properties and limitations. Section 4 outlines our methodology for constructing an adversarial framework designed to manipulate PD plots. Section 5 examines the effectiveness of our framework using real-world datasets. Section 6 offers insights and recommendations for regulators and practitioners. Section 7 concludes the paper.


\section{Related Studies}


There are very few studies on adversarial attacks on global interpretation methods applied to tabular data, even though these methods are becoming increasingly popular in critical areas such as insurance and credit scoring. In a related work, \citet{baniecki2023fooling} approached fooling PD as an optimization problem using genetic and gradient algorithms, with adversarial data perturbations by supplying a poisoned dataset. Most adversarial attacks on interpretation methods involve either perturbing the data through small changes or modifying the model itself. Our adversarial framework can be viewed as a modification of the model and is distinctive for preserving the original model's predictive performance while enabling auditors to examine the input dataset.

In a notable earlier study, \citet{slack2020fooling} demonstrated that local interpretation methods relying on input perturbations, such as LIME \citep{ribeiro2016should} and SHAP \citep{lundberg2017unified}, can be vulnerable to adversarial attacks. These attacks exploit the use of synthetic neighborhood data points generated by LIME or SHAP, deviating from the underlying data distribution. Building upon this notion, our work expands on this idea by identifying and exploiting the extrapolated data generated by PD to deceive the PD itself via an adversarial framework. Our approach can be seen as an extension of \citet{slack2020fooling}'s scaffolded classifier to a scaffolded regressor framework, which serves as the inspiration for our methodology. Apart from these, other global or local interpretation methods for tabular data are susceptible to adversarial attacks, including global SHAP \citep{baniecki2022manipulating, laberge2022fooling} and counterfactual explanations \citep{slack2021counterfactual}.

In the broader landscape of machine learning, the existing literature on adversarial attacks of interpretation tools extensively covers attacks targeting model-specific explanations. These studies primarily focus on deep neural networks applied to image data, as evidenced by works such as \citet{ghorbani2019interpretation, dombrowski2019explanations, heo2019fooling, dimanov2020you}, rather than the context of tabular data that we explore. Prior to these works, earlier studies primarily focused on adversarial attacks targeting model predictions \citep{szegedy2013intriguing, goodfellow2014explaining}. Interestingly, adversarial examples were generated to deceive model classifications in \citet{goodfellow2014explaining} by altering pixel values to fall outside the training data distribution. This exploits the vulnerabilities of certain neural network architectures to ``extrapolation'' in image classification for conducting adversarial attacks. In comparison, we exploit the extrapolation vulnerabilities of PD plots -- in a different context for permutation-based interpretation methods -- to carry out adversarial attacks. For a comprehensive survey on adversarial attacks on model explanations, along with the corresponding defenses against such attacks, we direct readers to \citet{baniecki2023adversarial}.

\section{Partial Dependence Plots}


\subsection{Notation and Definitions}

Let $\mathrm{y}$ be the target variable, $\textbf{x}$ be a set of predictor variables, and $\hat{f}$ be a trained predictive machine learning (ML) model, $\hat{\mathrm{y}} = \hat{f}(\textbf{x})$. The feature matrix is denoted by $X$, and the target vector is $\textbf{y}$. In this paper, the superscript $(i)$ indicates an individual observation, and the subscript $j$ refers to a specific featu: $x^{(i)}_j$ specifies the $j$th feature of the $i$th observation. Given the observed training data $\{X,\textbf{y}\} = \{\textbf{x}^{(i)},  \mathrm{y}^{(i)}\}^n_{i=1}$, $y^{(i)} \in \mathbb{R}$ and $\textbf{x}^{(i)} \in \mathbb{R}^p$, where $n \in \mathbb{N}$ is the number of observations and $p \in \mathbb{N}$ is the number of features.


\textbf{Partial Dependence (PD) Plot:} The PD plot displays the marginal effect of a feature on the prediction \citep{ friedman2001greedy}. Let $X_S$ be a feature subset of $X$ (typically $|S| = 1 \; \text{or} \; 2$), denoted as $X_S \subset \{X_1,\dots,X_p\}$, and $X_C$ be the complement subset, $X_S \cup X_C = X$. The PD function is defined as
\begin{align} \label{eqn:pdpS}
\mathrm{PD}_S (X_S)  = \mathbb{E}_{X_C}[\hat{f}(X_S, X_C)] = \int \hat{f}(X_S, X_C) d\mathbb{P}(X_C)
\end{align}
When there is only one feature of interest $X_j$ in $S$, let $\textbf{x}^{(i)}_{-j}$ denote the $i^{th}$ row of $X$ without the $j^{th}$ feature, the above can be estimated from the training sample by
\begin{equation} \label{eqn:pdp}
    \mathrm{\widehat{PD}}_j (x_j) = \frac{1}{n} \sum_{i=1}^n \hat{f}(x_j, \textbf{x}^{(i)}_{-j})
\end{equation}


The PD function calculates the average predicted outcome when the $j$th column of $X$ is replaced with the value $x_j$. It is important to note that PD plots assume that the features in $X_S$ are independent of the features in $X_C$. In general, PD plots provide a global model-agnostic technique that reveals the global relationship between a feature and the predicted output.

\textbf{Marginal plots (M-plots)} \citep{apley2020visualizing} provide an alternative to PD plots by focusing on the conditional distribution, effectively addressing extrapolation issues that may arise from correlated features. The function of a marginal plot is defined as
\begin{align} \label{eqn:mplot}
\mathrm{M}_S  = \mathbb{E}_{X_C|X_S}[\hat{f}(X_S,X_C)|X_S=x_S] = \int \hat{f}(X_S, X_C) d\mathbb{P}(X_C|X_S = x_S)
\end{align}
However, a limitation of M-plots arises when a feature of interest, $j$, is correlated with an unplotted feature $k$ in $X$. In such cases, the M-plot for feature $j$ show a mixed effect of both features $j$ and $k$, even if feature $j$ has no predictive power on the target variable. For this reason, M-plots have limited utility as tools for assessing main effects \citep{friedman2001greedy,apley2020visualizing,gromping2020model}.


\subsection{Properties of PD Plots}

\textbf{Additive or Multiplicative Recovery Properties:} If the dependence of $\hat{f}(X)$ on $X_S$ is additive: $\hat{f}(X) = \hat{h}_S(X_S) + \hat{h}_C(X_C)$, then $\mathrm{PD}_S (X_S)$ is equal to $\hat{h}_S(X_S)$, up to an additive constant. If the dependence of $\hat{f}(X)$ on $X_S$ is multiplicative: $\hat{f}(X) = \hat{h}_S(X_S)  \cdot \hat{h}_C(X_C)$, then $\mathrm{PD}_S (X_S)$ is equal to $\hat{h}_S(X_S)$, up to an multiplicative constant factor \citep{friedman2001greedy,hastie2009elements}. Marginal plots, in comparison, do not conform to these properties. \citet{gromping2020model} argued that PD plots are more conceptually sound than M-plots or Accumulated Local Effect (ALE) plots -- a popular alternative to PD plots.

\textbf{Causal interpretations}:  It is possible to derive causal interpretations from black-box models using PD plots: PD plots estimate the causal effect of $X_S$ on $Y$, provided that $X_C$ satisfies the back-door criterion \citep{zhao2021causal}. \citet{loftus2023causal} proposed causal dependence plots -- a causal analog of PD plot, based on the structural causal model, however, it can be a challenge to identify the structural causal model in real-world applications.

\subsection{Limitations of PD Plots}

\textbf{Extrapolation:} PD plots, along with other perturbation-based interpretation methods, does not account for the interaction effects among features. Consequently, when there is strong dependence among features, these methods may yield misleading interpretation results as they extrapolate to regions with little or no data \citep{hooker2004diagnosing,hooker2021unrestricted}. Furthermore, the selection of sampling points, such as the use of equidistant grids, can exacerbate the problem of extrapolation \citep{molnar2022general,krause2016interacting}.

\textbf{Aggregation of Heterogeneous Effects:} PD plots represent an average curve, but they can obscure opposite behaviors present within subsets of the data. For example, a flat PDP curve may suggest one of two scenarios: (1) the feature has no significant influence on the prediction, or (2) different subsets of the dataset exhibit opposing trends that offset each other when calculating the overall average \citep{angelini2023visual,molnar2022general}.

\textbf{Ignorance of Uncertainty:} PD plot fails to consider the uncertainty in both its estimation and model fitting. For instance, its interpretations can be misleading in situations where models are underfitted or overfitted \citep{molnar2022general}. Instead, the uncertainty of PD plots can be estimated on bootstrap samples for a given model to construct confidence intervals \citep{cafri2016understanding}.

Crucially, the first two limitations -- the extrapolation behavior of correlated features and the aggregation of heterogeneous effects during the averaging process -- are exploited in Section \ref{fooling} to manipulate PD plot outputs.

\section{The Process of Fooling Partial Dependence Plots} \label{fooling}

In this section, we outline our methodology for constructing an adversarial framework, denoted by $a(\textbf{x})$, which is designed to replace the original ML model $f(\textbf{x})$.  Initially, we consider a simpler scenario of manipulating a single PD plot. Building on this foundation, we expand our methodology to simultaneously fool multiple PD plots within a unified framework.

\subsection{Adversarial Framework a(\textbf{x})}

\begin{figure}[h]
  \centering
  \includegraphics[width=0.9\textwidth]{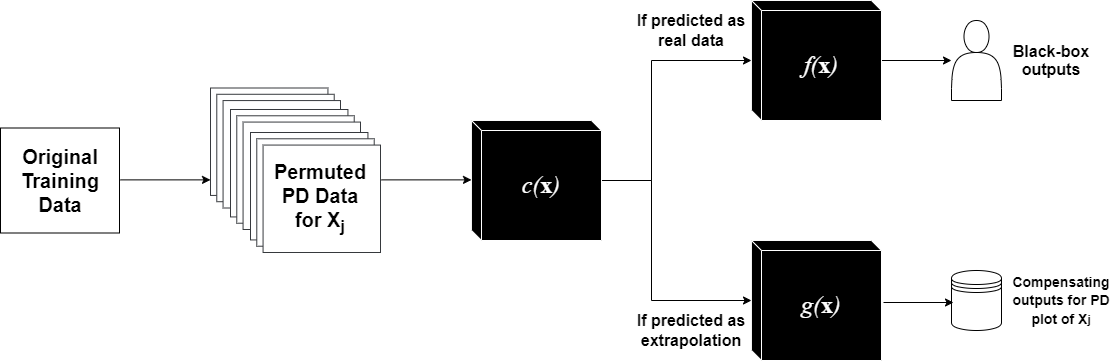}
  \caption{Adversarial Framework \( a(\textbf{x}) \) for Manipulating the PD Plot of Feature \( X_j \)}
  \label{fig:ax_onevar}
\end{figure}

Our adversarial framework $a(\textbf{x})$ takes the form:
\begin{align} \label{eqn:ax}
a(\textbf{x}) = (1- c(\textbf{x})) \cdot f(\textbf{x}) + c(\textbf{x}) \cdot g(\textbf{x}).
\end{align}
The model $a(\textbf{x})$ replaces the original machine learning model $f(\textbf{x})$, which aims to maintain the prediction accuracy of the original model and generate manipulated PD plots.

Here, $c(\textbf{x})$ is a classifier to distinguish between instances from the original feature space and those from the extrapolation domain. $g (\textbf{x})$ is a function to provide specific outputs to form the desired PD plots. The framework predicts that an instance originates from the original feature space when $c(\textbf{x}) = 0$ and outputs the black-box prediction $f(\textbf{x})$ accordingly. Conversely, when $c(\textbf{x}) = 1$, the instance is determined to come from the extrapolation domain, and $a(\textbf{x})$ outputs $g(\textbf{x})$ -- the compensating outputs. Broadly, the framework uses $g(\mathbf{x})$ for instances in the extrapolation domain as identified by $c(\mathbf{x})$, aiming to compensate for the discriminatory performance of model $f(\textbf{x})$ on real data when generating PD plot outputs. A simple scenario of this process is visually summarized in Figure \ref{fig:ax_onevar}.

The manipulated PD function for the $j^{th}$ feature after adversarial attacks can be expressed as:
\begin{align}
\mathrm{\widehat{PD}}_j^\text{adv}(x_j) &= \frac{1}{n} \sum_{i=1}^n \hat{a}(x_j, \textbf{x}^{(i)}_{-j}) \\
\label{eqn:pdp_adv1}
&= \frac{1}{n} \left(\sum_{i=1}^n \hat{f}(x_j, \textbf{x}^{(i)}_{-j}) \cdot (1 - \hat{c}(x_j, \textbf{x}^{(i)}_{-j})) \right. + \left. \sum_{i=1}^n \hat{g}(x_j, \textbf{x}^{(i)}_{-j}) \cdot \hat{c}(x_j, \textbf{x}^{(i)}_{-j}) \right)
\end{align}
We simplify this expression to:
\begin{align}
\mathrm{\widehat{PD}}_j^\text{adv}(x_j) = (1-\hat{\lambda}_j(x_j)) \hat{\rho}_j(x_j) + \hat{\lambda}_j(x_j) \hat{\gamma}_j(x_j), \label{eqn:pdadv}
\end{align}
where $\lambda_j(x_j)$ represents the proportion of permuted data identified as belonging to the extrapolation domain by $\hat{c}(\textbf{x})$ for feature $j$ at value $x_j$. The term $\rho_j(x_j)$ is the conditional PD function, which serves as an approximation of $\mathrm{M}_j(x_j)$ in (\ref{eqn:mplot}), capturing the global relationship between the feature and the predicted output, considering non-extrapolation data only. $\gamma_j(x_j)$ denote the compensating output from model $g$ for feature $X_j$ at value $x_j$, when attempting to fool one PD plot at a time. We have $g(x_j, \textbf{x}^{(i)}_{-j}) = \gamma_j(x_j)$. These components are estimated as follows:
\begin{align}
\hat{\lambda}_j(x_j) & = \frac{1}{n} \sum_{i=1}^{n}  \hat{c}(x_j, \textbf{x}^{(i)}_{-j}) \\
\hat{\rho}_j(x_j) &=  \frac{1}{n-n \hat{\lambda}_j(x_j)} \sum_{i=1}^n \hat{f}(x_j, \textbf{x}^{(i)}_{-j}) \cdot (1 - \hat{c}(x_j, \textbf{x}^{(i)}_{-j}))
\end{align}
Given these, if $\mathrm{\overline {PD}}_j^\text{adv}(x_j)$ denotes the desired PD output set by modelers (attackers) for feature $X_j$ at value $x_j$ we derive:
\begin{align} \label{eqn:compop}
 g(x_j, \textbf{x}^{(i)}_{-j})=\hat{\gamma}_j(x_j) =  \frac{\mathrm{\overline {PD}}_j^\text{adv}(x_j) - (1-\hat{\lambda}_j(x_j)) \hat{\rho}_j(x_j)}{\hat{\lambda}_j(x_j)}.
\end{align}

In the following subsections, we discuss how to construct $c(\textbf{x})$ and $g(\textbf{x})$ and apply the adversarial framework to fool single and multiple PD plots.

\subsection{Construct the Classifier $c(\textbf{x})$}
\label{sec: cx}
To build the adversarial framework, we commence with the construction of classifier $c(\textbf{x})$ to distinguish between instances from the original feature space and those from the extrapolation domain. To construct such a classifier, we first generate an augmenting sample $\tilde{X} = \{\tilde{\textbf{x}}^{(i)}\}^{\tilde{n}}_{i=1}$ following the approach proposed by \citet{hooker2012prediction}, complementing the training of $c(\textbf{x})$ with original data $X$. Let $\mathbb{P}(\tilde{\textbf{x}})$ denote the joint (generative) feature distribution to generate $\tilde{X}$, and set it as a uniform distribution based on the empirical range of the training data:
\begin{align}
     \mathbb{P}(\tilde{\textbf{x}}) &= \mathbb{P}(\tilde{\textbf{x}}; X) = \prod_{j=1}^p \mathcal{U}(\min_i(x^{(i)}_j) < \tilde{x}_j < \max_i(x^{(i)}_j)),
\end{align}
where \(\mathcal{U}(a, b)\) represents a uniform marginal distribution ranging from \(a\) to \(b\), and $\min(x^{(i)}_j)$ and $\max(x^{(i)}_j)$ denote the minimum and maximum values of the $j$th feature in the training sample. In other words, each feature $\tilde{X}_j$ in $\tilde{X}$ is independently generated from the uniform marginal distribution of individual features.

The proposal of uniform $\mathbb{P}(\tilde{\textbf{x}})$ was introduced by \citet{hooker2012prediction} for a different objective, aiming to provide regularization in regions of sparse data to control the model's extrapolation behavior and improve its prediction performance. In our study, the augmenting sample $\tilde{X}$ is drawn randomly and independently among features according to the uniform $\mathbb{P}(\tilde{\textbf{x}})$. This process disrupts the dependence among features in the augmenting sample, leading to extrapolation. Subsequently, we combine the real data and augmenting (uniform) data, denoted as $X \cup \tilde{X}$, and assign labels $c(\textbf{x}) = 0$ to instances in $X$ and $c(\textbf{x}) = 1$ to instances in $\tilde{X}$. The classifier $c(\textbf{x})$ is trained using the combined dataset $X \cup \tilde{X}$ along with their corresponding labels to differentiate between real and uniform data. This approach enables $c(\textbf{x})$ to learn the interrelationships within the real data $X$, thus effectively distinguishing instances that originate from the feature space defined by the training set from those in the extrapolation domain.


Our classifier $c(\textbf{x})$ shares similarities with the out-of-distribution (OOD) classifier in \citet{slack2020fooling}, which directly used the permuted data generated by LIME or kernel SHAP as their augmenting data. However, our approach diverges from directly using permuted data used in generating PD plots to construct $c(\textbf{x})$ for the following reasons: firstly, we do not treat all permuted PD data as instances of extrapolation\footnote{We believe that the effective fooling of LIME (as compared to kernel SHAP) as observed in \citet{slack2020fooling} can be partially attributed to the fact that some permuted data used by LIME involve extrapolation within the training data range (e.g., if the feature `age' consists only of integer values in all training data, while the permuted data include non-integer ages). This occurs because LIME is implemented by randomly perturbing the feature values of the original instance, which explains why directly using the permuted data works in Slack's approach. However, the fooling effectiveness of our algorithm does not benefit from this aspect (i.e., the term ``permutation'' does not refer to exactly the same thing in LIME and PD plots). For more details on the implementation of our algorithm, refer to Appendix \ref{sec:more_cx}. \label{fnote1}}; secondly, our adversarial framework $a(\textbf{x})$ uses only extrapolated permuted PD data as compensating outputs, rather than using all permuted PD data, thereby enhancing the efficiency and accuracy of the fooling process.





\subsection{Apply the Adversarial Framework to Fool One PD Plot} \label{sec: fool onepd}

Now we have constructed our adversarial framework in Equation \ref{eqn:ax} after constructing $c(\textbf{x})$ and $g(\textbf{x})$ in previous subsections. In this subsection, we introduce the process of implementing the adversarial attack to fool one PD plot at a time.

Let us denote the set of grid values which we will use to plot $\widehat{PD}_j(x_j)$ by $V_j = \{v_j^1, v_j^2, ..., v_j^{m_j}\}$, where $m_j$ is the number of grid points and each $v_j^p$ ($p\in\{1,2,\dots, m_j\}$) represents a specific grid value in $V_j$. For each $v_j^p \in V_j$, we create a permuted dataset by replacing the $j$th feature in all training data points with $v_j^p$. The set of all permuted datasets is then represented as:
   \[
   \mathcal{P}_j = \left\{ \left( \mathbf{x}^{(i)}_{[j \to v_j^p]}, y^{(i)} \right) \right\}_{i=1, p=1}^{n, m_j}
   \]
Here, $\mathbf{x}^{(i)}_{[j \to v_j^p]}$ represents the feature vector of the $i$ th instance with the $j$th feature replaced by $v_j^p$, done for all $n$ instances and all $m$ grid values. For the experimental results presented in Section \ref{sec:exper_results}, we carefully select grid values that are directly drawn from the training dataset, ensuring that each $V_j$ is an observed value, thereby avoiding extrapolation within the range of the training data. This approach ensures that our algorithm's effectiveness is not artificially enhanced by extrapolated grid values. For further discussions and implementation details on this aspect, please refer to footnote \ref{fnote1} and Appendix \ref{sec:more_cx}.

\begin{algorithm}[h!]
\caption{Fooling One Partial Dependence Plot} \label{alg:fooling_onepd}
\begin{algorithmic}[1]
\Statex \textbf{Input:} Training data $X$, feature of interest $X_j$, grid values $V_j$, permuted PD data $\mathcal{P}_j$, target PD outputs $\overline{\mathrm{PD}}_j^\text{adv}$, trained ML model $\hat{f}(\mathbf{x})$.
\Statex \textbf{Output:} Manipulated PD outputs $\mathrm{\widehat{PD}}_j^\text{adv}(v_j^p)$ for each $v_j^p \in V_j$

\State Generate augmenting sample $\tilde{X}$
\State Label each $\mathbf{x} \in X \cup \tilde{X}$ with $l_c = 0$ if $\mathbf{x} \in X$, and $l_c = 1$ if $\mathbf{x} \in \tilde{X}$.
\State Train classifier $c(\mathbf{x})$ using $D_c = \{(\mathbf{x}, l_c) \mid \mathbf{x} \in X \cup \tilde{X}, l_c \in \{0, 1\}\}$.
\For{each $v_j^p \in V_j$}
    \State Estimate $\hat{\lambda}_j(v_j^p)$ using $\mathcal{P}_j$ and $\hat{c}(\mathbf{x})$
    \State Estimate $\hat{\rho}_j(v_j^p)$ using $\mathcal{P}_j$, $\hat{c}(\mathbf{x})$ and $\hat{f}(\mathbf{x})$
    \State Estimate $\hat{\gamma}_j(v_j^p)$ using $\mathrm{\overline{PD}}_j^\text{adv}(v_j^p)$, $\hat{\lambda}_j(v_j^p)$ and $\hat{\rho}_j(v_j^p)$
\EndFor
\For{each $v_j^p \in V_j$}
    \State Initialize an empty list $A$ to store $\hat{a}(\textbf{x})$ predictions at value $v_j^p$
    \For{each $\mathbf{x} \in \mathcal{P}_j$ where $x^{(i)}_j = v_j^p$}
        \State $\hat{a}(\mathbf{x}) = \begin{cases}
          \hat{f}(\mathbf{x}) & \text{if } \hat{c}(\mathbf{x}) = 0 \\
          \hat{g}(\mathbf{x}) = \hat{\gamma}_j(v_j^p) & \text{if } \hat{c}(\mathbf{x}) = 1
          \end{cases}$
        \State Add $\hat{a}(\mathbf{x})$ to $A$
    \EndFor
    \State Output $\mathrm{\widehat{PD}}_j^\text{adv}(v_j^p)$ = average of $A$
\EndFor
\end{algorithmic}
\end{algorithm}

The process for manipulating a single PD plot is summarized in Algorithm \ref{alg:fooling_onepd}. For each permuted observation where the \(j\)th feature is set to a specific value, the model $\hat{c}(\mathbf{x})$ determines if it originates from real data or extrapolation domain. If $\hat{c}(\mathbf{x})$ predicts the observation as real data, the output will remain as the black-box model's output $f(\textbf{x})$; if it is predicted as extrapolation, the output switches to a predetermined value $\hat{\gamma}_j(x_j)$, based on the value of feature \( j \). This procedure is formulated in (\ref{eqn:ax}) and illustrated in Figure \ref{fig:ax_onevar}. The same process applies when making predictions on the test data, and the goal is for our adversarial algorithm \( a(\textbf{x}) \) to retain as many predictions of $f(\textbf{x})$ as possible.

\subsection{Construct $g(\textbf{x})$ to Fool Multiple PD Plots} \label{sec: gx fooling_multipd}

In practice, modelers may wish to fool multiple PD plots. However, independent manipulation of each PD plot could lead to inconsistent model predictions: if an observation is identified by $\hat{c}(\textbf{x})$ as extrapolation, we need to select which feature's compensating outputs to use. To address this, we introduce a unified $g(\textbf{x})$ model designed to ensure consistent prediction outputs while manipulating $q$ PD plots simultaneously, where $1 < q \leq p$. Unlike the case of manipulating a single PD plot, we augment $g$ with a specialized classifier $c_1(\textbf{x})$, which is trained with $q+1$ class responses to allocate the appropriate compensating outputs to each PD plot accurately. The set of all potential classes for $c_1(\textbf{x})$ is denoted as $G = \{G_1, G_2, \dots, G_q, G_{\text{no}}\}$, including a base class $G_{\text{no}}$ for real instances and classes $G_1$ to $G_q$ corresponding to each of the $q$ features targeted for manipulation. To train $c_1(\textbf{x})$ we use the real data $X$ along with the permuted PD data for the $q$ targeted features identified as extrapolation by $\hat{c}(\textbf{x})$, where target labels are given by which feature (or none) were permuted in producing each datum.


For clarity, we use $k$ to denote the predicted index of the classifier $c_1(\textbf{x})$ among the $q+1$ classes, distinct from index $j$ for all $p$ features in $X$, such that $\hat{c}_1(\mathbf{x}) = G_k, G_k \in G$. For a permuted observation $g(x_j, \textbf{x}^{(i)}_{-j})$ classified by $c_1(\textbf{x})$ as belonging to class $G_k$, the output of model $g$ is then given by
\begin{align}
g(\textbf{x}) = \left\{
\begin{array}{lcl}
   \gamma_k(x_k)  & & \text{if} \enspace \hat{c}_1(\mathbf{x}) = G_k\\
   f(\textbf{x}) & & \text{if} \enspace \hat{c}_1(\mathbf{x}) = G_{\text{no}}.
\end{array}
\right.
\end{align}
We thus divide the extrapolation part of feature space into regions associated with permuting each feature of interest. We anticipate that permuting different features leads to little overlap in the extrapolation region and can form adversarial function $\gamma_k$ independently.  Note that our framework has a two-step classifier: first applying $c(\textbf{x})$ and then $c_1(\textbf{x})$.  This is to ensure that $f(\textbf{x})$ remains unperturbed on the original data distribution -- we will revert to the original model if either classifier tells us to.
The process of fooling two plots is illustrated in Figure \ref{fig:ax_twovar}. Notably, in the scenario of fooling multiple PD plots, only $c(\textbf{x})$ continues to affect the accuracy of $a(\textbf{x})$ -- the predictions retained of $a(\textbf{x})$ from $f(\textbf{x})$. The inclusion of $c_1(\textbf{x})$ only affects the stability or uncertainty of $a(\textbf{x})$ (how $\mathrm{\widehat{PD}}_j^\text{adv}(x_j)$ deviates from $\mathrm{\overline {PD}}_j^\text{adv}(x_j)$).

\subsection{Apply the Adversarial Framework to Fool Multiple PD Plots} \label{sec: fool multipd}

\begin{figure}[h]
  \centering
  \includegraphics[width=0.9\textwidth]{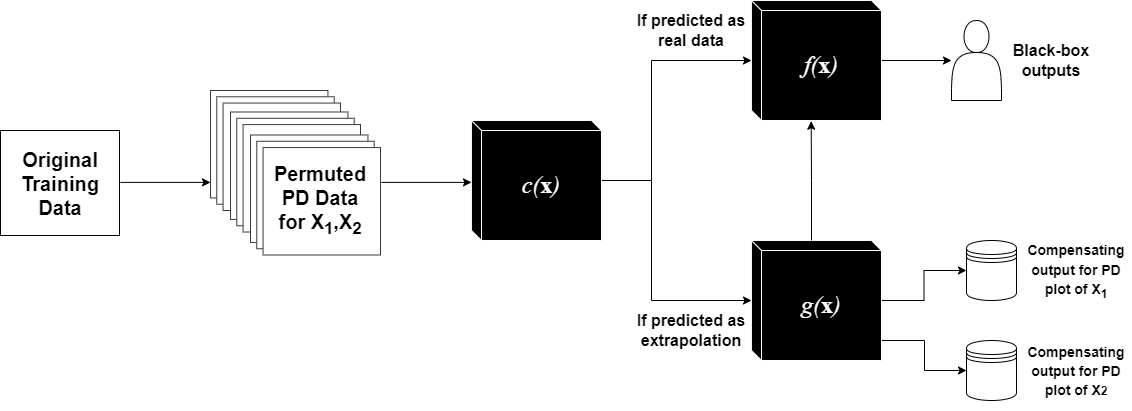}
  \caption{Adversarial Framework \( a(\textbf{x}) \) for Manipulating the PD Plots of Features \( X_1\) and \( X_2\)}
  \label{fig:ax_twovar}
\end{figure}

As summarized in Algorithm \ref{alg:fooling_onepd}, let $X_k$ represent an additional targeted feature for manipulation. We define $V_k = \{v_k^1, v_k^2, ..., v_k^{m_k}\}$ as the set of grid values corresponding to this feature. The process for manipulating two PD plots within a unified $g(\textbf{x})$ model is outlined in Algorithm \ref{alg:fooling_twopd} and depicted in Figure \ref{fig:ax_twovar}, noting that our framework can be readily adapted for fooling multiple PD plots.
In comparison, Algorithm \ref{alg:fooling_twopd} leverages compensating outputs derived from Algorithm \ref{alg:fooling_onepd}, and the integration of $c_1(\mathbf{x})$ in Algorithm \ref{alg:fooling_twopd} introduces additional variability into the fooling process compared to Algorithm \ref{alg:fooling_onepd}.

\begin{algorithm}[h!]
\caption{Fooling Two Partial Dependence Plots Using a Single $g(\mathbf{x})$ Model} \label{alg:fooling_twopd}
\begin{algorithmic}[1]
\Statex \textbf{Input:} Training data $X$, features of interest $X_j, X_k$, grid values $V_j, V_k$, permuted PD data $\mathcal{P}_j, \mathcal{P}_k$, target PD outputs $\overline{\mathrm{PD}}_j^\text{adv}, \overline{\mathrm{PD}}_k^\text{adv}$, trained ML model $\hat{f}(\mathbf{x})$.
\Statex \textbf{Output:} Manipulated PD outputs $\mathrm{\widehat{PD}}_j^\text{adv}(v_j^p)$ for each $v_j^p \in V_j$, $\mathrm{\widehat{PD}}_k^\text{adv}(v_k^p)$ for each $v_k^p \in V_k$

\State Perform steps 1-7 from Algorithm 1 for both features $X_j$ and $X_k$ to estimate $\hat{\lambda}_j, \hat{\rho}_j, \hat{\gamma}_j$ for each $v_j^p \in V_j$ and $\hat{\lambda}_k, \hat{\rho}_k, \hat{\gamma}_k$ for each $v_k^p \in V_k$.
\State Label each $\mathbf{x} \in X \, \lor \, (\mathbf{x} \in \mathcal{P}_j \cup \mathcal{P}_k \, \land \, \hat{c}(\mathbf{x}) = 1)$ with $l_g = 0$ if $x \in X$, $l_g = 1$ if $x \in \mathcal{P}_j \, \land \, \hat{c}(\mathbf{x}) = 1$, and $l_g = 2$ if $x \in \mathcal{P}_k \, \land \, \hat{c}(\mathbf{x}) = 1$.
\State Train classifier $c_1(\mathbf{x})$ using $D_g = \{(\mathbf{x}, l_g) \mid \mathbf{x} \in X \, \lor \, (\mathbf{x} \in \mathcal{P}_j \cup \mathcal{P}_k \, \land \, \hat{c}(\mathbf{x}) = 1), l_g \in \{0, 1, 2\}\}$.
\For{each $v_j^p \in V_j$}
    \State Initialize an empty list $A$ to store $\hat{a}(\textbf{x})$ predictions at value $v_j^p$
    \For{each $\mathbf{x} \in \mathcal{P}_j$ where $x^{(i)}_j = v_j^p$}
        \State $\hat{a}(\mathbf{x}) = \begin{cases}
          \hat{f}(\mathbf{x}) & \text{if } \hat{c}(\mathbf{x}) = 0 \, \lor \, (\hat{c}(\mathbf{x}) = 1 \land \hat{c}_1(\mathbf{x}) = 0) \\
          \hat{g}(\mathbf{x}) = \hat{\gamma}_j(v_j^p) & \text{if } \hat{c}(\mathbf{x}) = 1 \, \land \, \hat{c}_1(\mathbf{x}) = 1 \\
          \hat{g}(\mathbf{x}) = \hat{\gamma}_k(x_k^{(i)}) & \text{if } \hat{c}(\mathbf{x}) = 1 \, \land \, \hat{c}_1(\mathbf{x}) = 2 \enspace \text{ (interpolate or extrapolate if needed)} \\
          \end{cases}$
        \State Add $\hat{a}(\mathbf{x})$ to $A$
    \EndFor
    \State Output $\mathrm{\widehat{PD}}_j^\text{adv}(v_j^p)$ = average of $A$
\EndFor
\State Perform steps 4-11 to compute and output manipulated PD outputs $\mathrm{\widehat{PD}}_k^\text{adv}(v_k^p)$ for each $v_k^p \in V_k$.
\Statex \textbf{Note:} When $\hat{c}_1(\mathbf{x})$ incorrectly identifies
extrapolation to another feature, the compensating output $g(\mathbf{x})$ may require interpolation or extrapolation for continuous features not directly included in $V_k$ (or $V_j$ when fooling $X_k$).
\end{algorithmic}
\end{algorithm}

In summary, our framework employs extrapolated predictions identified by $c(\textbf{x})$ to compensate for the discriminatory performance of model $f(\textbf{x})$ on real data. It maintains the performance of $f(\textbf{x})$ while concealing biases in the predictions on real data, rendering the framework seemingly neutral when interpreting the results through PD plots. Conceptually, we can regard the entire process as a unified black-box model, even though it involves the sequential construction of separate models. Instead of aggregating all model outputs as in (\ref{eqn:ax}), an alternative approach is to retrain \(f(\textbf{x})\) incorporating extrapolated permuted instances with their corresponding compensating outputs in the final step, but we note that this yields less fine control over the resulting predictions and partial dependence plots.

\section{Experimental Results} \label{sec:exper_results}

\subsection{Insurance Data}

This study evaluates the performance of our algorithm on real-world insurance datasets. We used the \texttt{pg17trainpol} and \texttt{pg17trainclaim} datasets, which were obtained from the R package CASdatasets \citep{dutang2020package} and used for the 2017 pricing game of the French institute of Actuaries. The data underwent preprocessing following the methodology proposed by \citet{havrylenko2022detection}. Using this insurance dataset, we construct a model to predict claim frequency, denoted as $f(\textbf{x})$. In this model, we use the number of claims (\texttt{claim\_nb}) as the target variable. We also selected 14 features as explanatory variables. For details of the variables and summary statistics of the dataset, please refer to Appendix \ref{sec:insurance_summary}.

\begin{figure}[h!]
  \centering
  \includegraphics[width=0.6\textwidth]{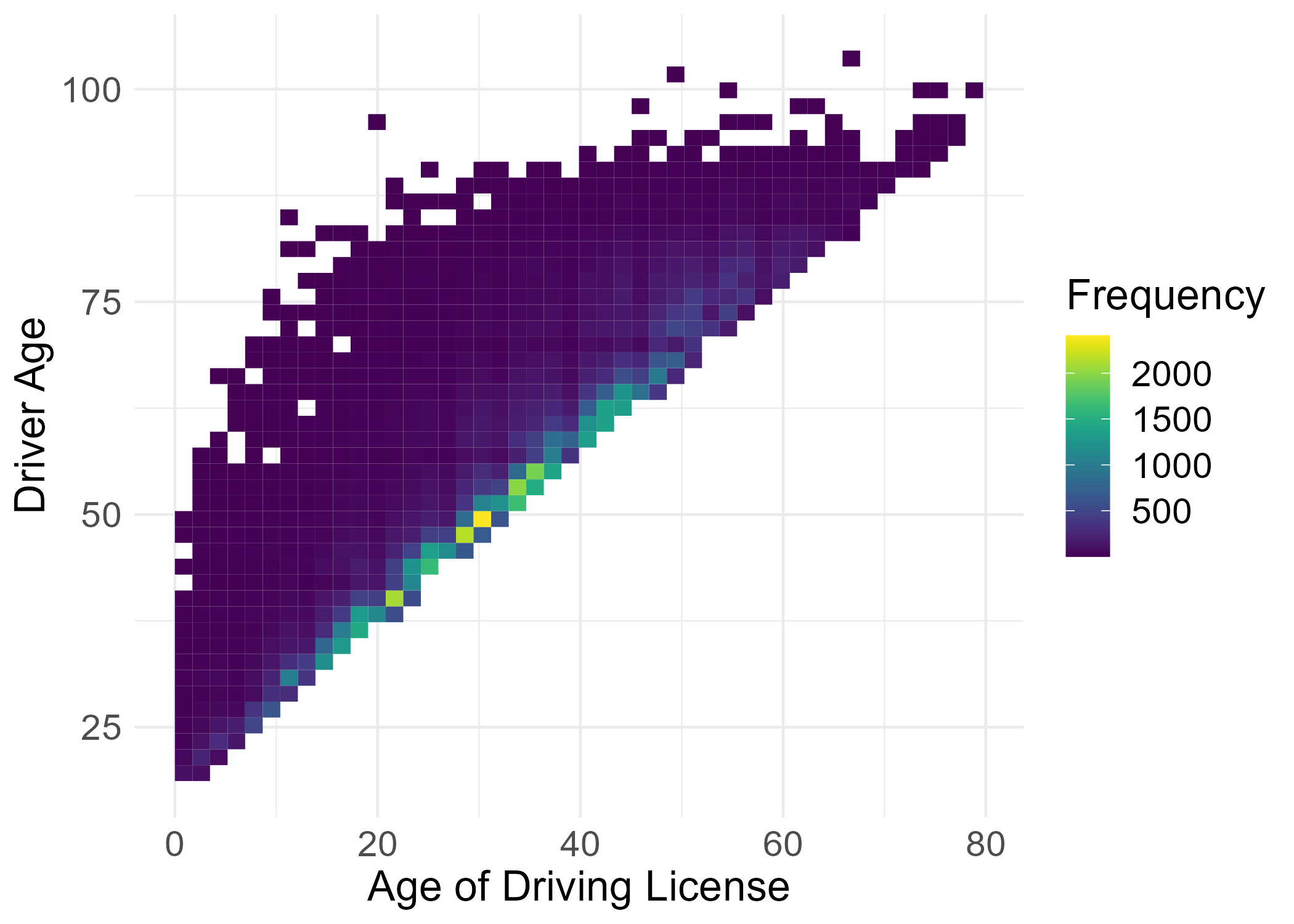}
  \caption{Relationship Between Driver's Age and Age of Driving License in the Insurance Dataset}
  \label{fig:drvage_drvlicage}
\end{figure}

Our study focuses on the driver's age (\texttt{drv\_age1}) as the first focal variable for adversarial attack, which is highly correlated with other time or experience-related variables, including a strong correlation with the age of the driving license (\texttt{drv\_age\_lic1}). Younger drivers typically possess a shorter driving history and therefore cannot be experienced, while older drivers are less likely to have learned to drive at a late age and are thus unlikely to be amateurs, as shown in Figure \ref{fig:drvage_drvlicage}.

In addition to exploring the driver's age, we also examine the vehicle value (\texttt{vh\_value}) as a secondary focal variable for adversarial attack, which is known to be correlated with other essential vehicle characteristics. This choice allows us to demonstrate that our algorithm effectively manipulates PD plots of multiple features simultaneously using Algorithm \ref{alg:fooling_twopd} (see Appendix \ref{sec:using_alg1} for results using Algorithm \ref{alg:fooling_onepd}). Additionally, the target PD outputs are set as a flat line at the average predictions, $\overline{\mathrm{PD}}_j^\text{adv} (v_j^p) = \hat{f}_{\text{avg}}(\textbf{x})$ for each $v_j^p \in V_j$. The choice of a flat target line is illustrative and can be easily extended to various targeted manipulations. For instance, we could perform a surgical-like manipulation on the PD plot for age, specifically targeting older ages, and only slightly adjust its PD values.

\begin{figure}[h!]
    \centering
    \includegraphics[width=\textwidth]{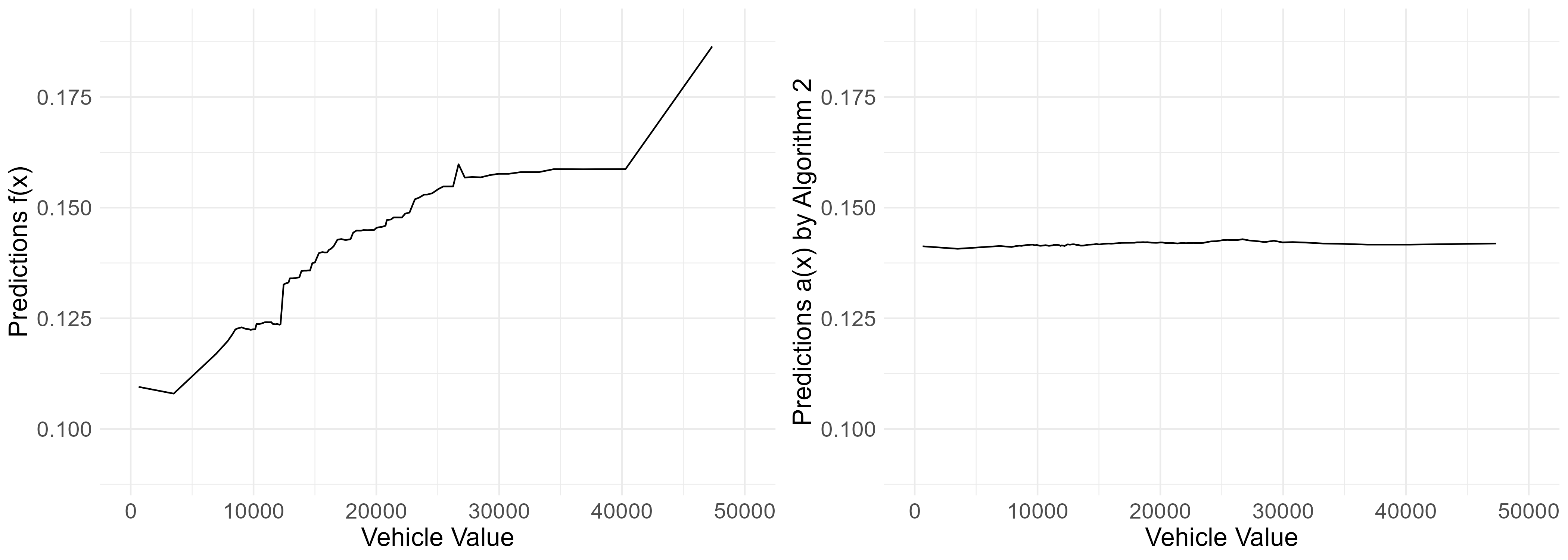}
    \caption{PD Plots for Vehicle Value before Attack (Left) and after Attack (Right), Insurance Data}
    \label{fig:pd_vhv_pg17}
\end{figure}

The classifier $c(\textbf{x})$ was constructed using the Keras package in R with a neural network architecture that comprises four fully connected layers with 40, 20, 10, and 1 units, respectively, and uses the rectified linear unit (ReLU) activation function. Likewise, the classifier $g(\textbf{x})$ was implemented with a similar neural network architecture but was trained based on permuted PD data. During the training of our neural networks, we employed techniques such as dropout and batch normalization to mitigate overfitting and enhance overall performance. The specifics of all neural network models trained in Section \ref{sec:exper_results} are detailed in Appendix \ref{sec:training_nn}. The black-box insurance model $f(\textbf{x})$ employed extreme gradient boosting (XGBoost) for claim frequency prediction. To optimize the performance of XGBoost, hyperparameters were carefully selected through a grid search approach with details provided in Appendix \ref{sec:training_xgboost}. Figures \ref{fig:pd_age_pg17} and \ref{fig:pd_vhv_pg17} showcase the resulting PD plots of the driver's age and the vehicle's value. It is important to note that the grid values for \texttt{vh\_value} were chosen based on the quantiles of the feature's distribution, rather than using equally spaced points. Notably, our adversarial attacks have proven to be effective, as is visually evident from the observed results.

For this evaluation, we divided the \texttt{pg17trainpol} and \texttt{pg17trainclaim} datasets into five folds. In each iteration, we assigned four folds of the data for training the adversarial framework and reserved the remaining one fold for testing purposes. The resulting PD plots of the driver's age and the vehicle's value calculated using the testing set are displayed in Figures \ref{fig:pd_age_pg17} and \ref{fig:pd_vhv_pg17}, respectively, under the condition that fold 1 is held out, and the threshold of $c(\textbf{x})$ is set to $0.5$. Our adversarial attacks have exhibited remarkable effectiveness, clearly evident from the visual analysis of the obtained results.

\begin{figure}[h!]
    \centering
    \includegraphics[width=\textwidth]{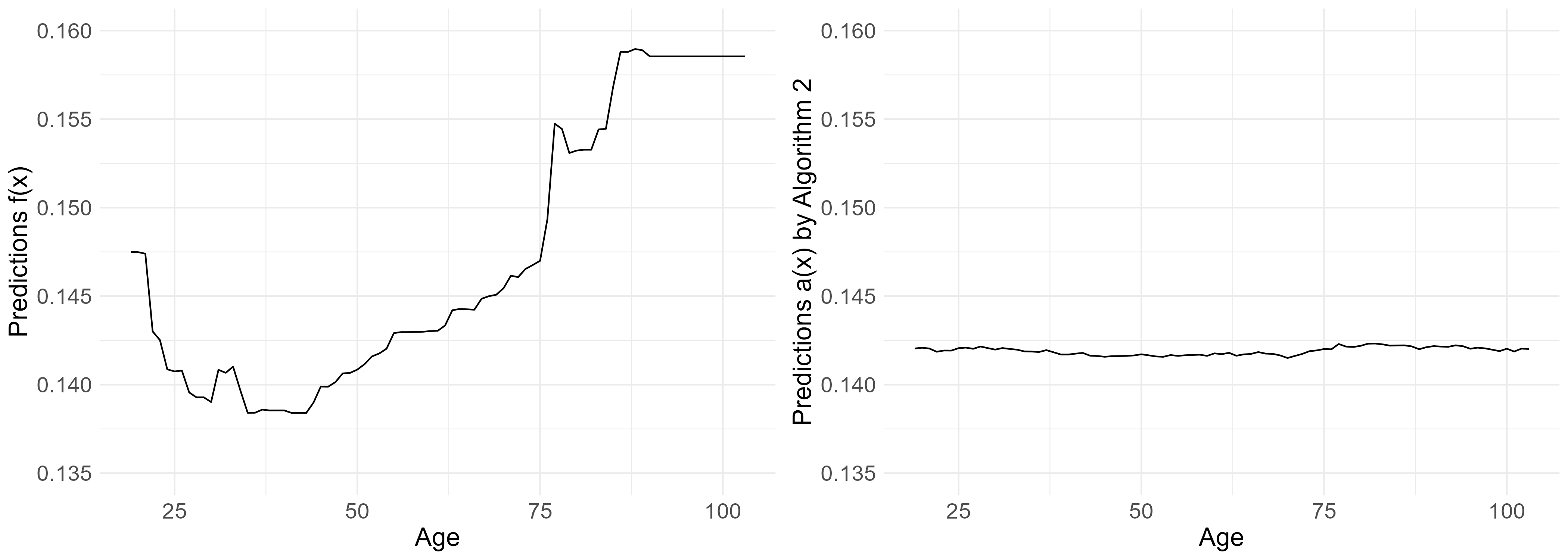}
    \caption{PD Plots for Age before Attack (Left) and after Attack (Right), Insurance Data}
    \label{fig:pd_age_pg17}
\end{figure}

\subsubsection*{Effect of Different Thresholds for c(\textbf{x})}

\begin{figure}[h!]
    \centering
    \includegraphics[width=\textwidth]{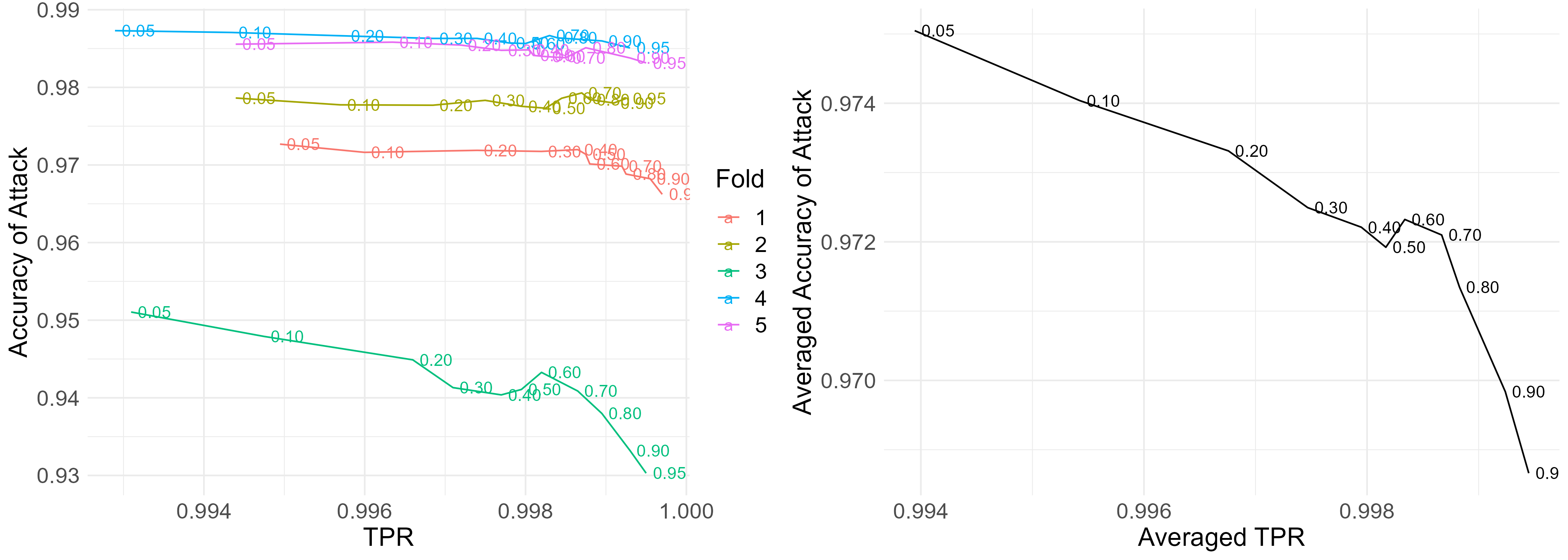}
    \caption{TPR versus Accuracy of Attack across Thresholds \(\{0.05, 0.1, 0.2, \ldots, 0.8, 0.9, 0.95\}\) (labeled next to the lines) for PD Plots of Age, Insurance Data}
    \label{fig:sd_age_pg17}
\end{figure}

\begin{figure}[h!]
    \centering
    \includegraphics[width=\textwidth]{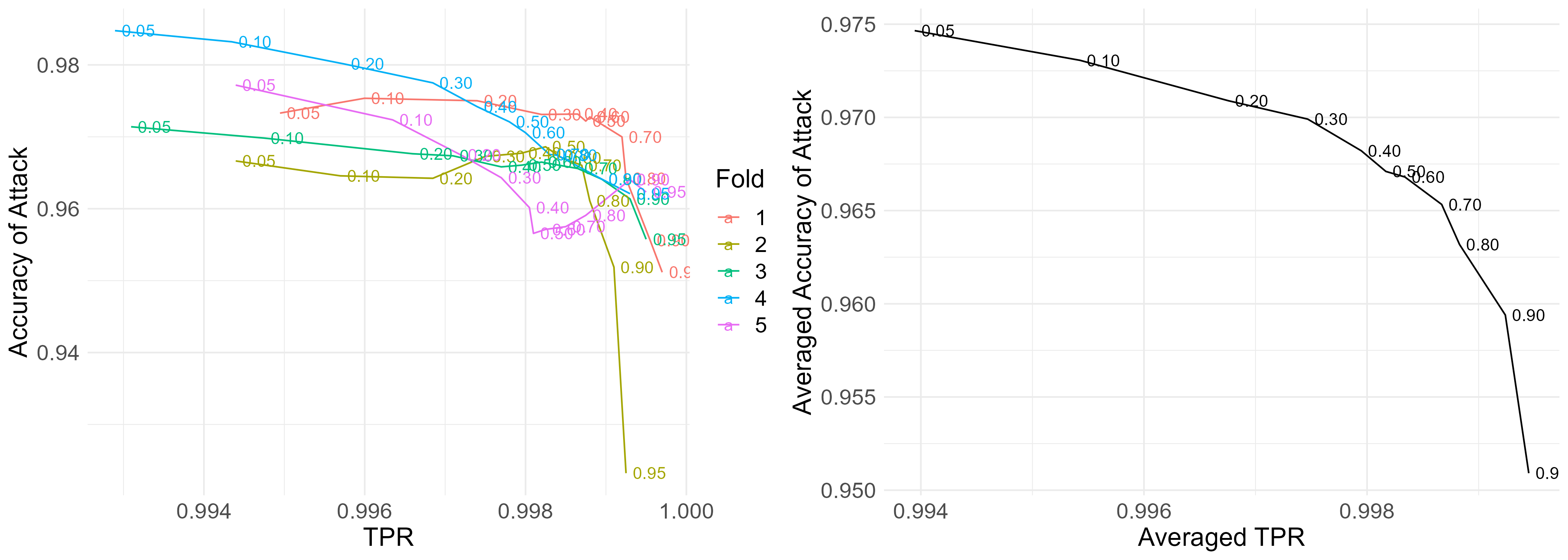}
    \caption{TPR versus Accuracy of Attack  across Thresholds \(\{0.05, 0.1, 0.2, \ldots, 0.8, 0.9, 0.95\}\) (labeled next to the lines) for PD Plots of Vehicle Value, Insurance Data}
    \label{fig:sd_vhv_pg17}
\end{figure}

In our adversarial framework, there exists an inevitable trade-off between the combined true positive rate of $c(\textbf{x})$ and $c_1(\textbf{x})$, which measures how closely $a(\textbf{x})$ retain the performance of $f(\textbf{x})$, and the accuracy of attack, measuring the effectiveness of adversarial attacks on PD plots. In our evaluation, we assess the impact of this trade-off on the hold-out test set by varying different thresholds for c(\textbf{x}). A lower threshold results in more data being labeled as extrapolation by $c(\textbf{x})$, enhancing the efficiency and accuracy of the fooling process but at the cost of reducing the proportion of $f(\textbf{x})$ retained by $a(\textbf{x})$. We define metrics for both true positive rate (TPR) and accuracy of attack as follows:
\begin{align} \label{eqn:acc_stab}
    \text{True Positive Rate (TPR)} &= \frac{1}{|X_{\text{test}}|} \sum_{\textbf{x} \in X_{\text{test}}} \left( \mathbb{I}(\hat{c}(\textbf{x}) = 0) + \mathbb{I}(\hat{c}(\textbf{x}) = 1 \land \hat{c}_1(\textbf{x}) = G_{\text{no}}) \right) \\
    \text{Accuracy of Attack} & = 1 - \frac{\|\mathrm{\widehat{PD}}_j^\text{adv}(v_j^p) - \mathrm{\overline {PD}}_j^\text{adv}(v_j^p)\|}{\| \mathrm{\widehat{PD}}_j^\text{orig}(v_j^p) - \mathrm{\overline {PD}}_j^\text{adv}(v_j^p)\|}
\end{align}
Here, $X_{\text{test}}$ refers to the hold-out test set. The true positive rate is defined as the proportion of data identified as non-extrapolation by either $c(\textbf{x})$ or $c_1(\textbf{x})$, indicating that the proportion of $f(\textbf{x})$ outputs are accurately preserved in $a(\textbf{x})$. The accuracy of the attack measures the average difference between the PD values estimated by the adversarial model and the desired PD values, scaled by the difference between the original PD values and the desired PD values. Higher values of these measures indicate higher proportion of f(\textbf{x}) retained in a(\textbf{x}) or more accurate fooling outcomes. The results, based on the tracks of five folds, are presented in Figures \ref{fig:sd_age_pg17} and \ref{fig:sd_vhv_pg17}. Notably, a generally decreasing trend is observed, highlighting the trade-off between TPR and accuracy of attack for both the PD plots of age and vehicle value. We attribute the increasing pattern, as seen in Figure \ref{fig:sd_age_pg17} when thresholds shift from 0.5 to 0.7, to the fact that age is less significant in modeling, resulting in less accurate manipulated PD plots.

\subsection{COMPAS Data}

\begin{figure}[h!]
    \centering
    \includegraphics[width=\textwidth]{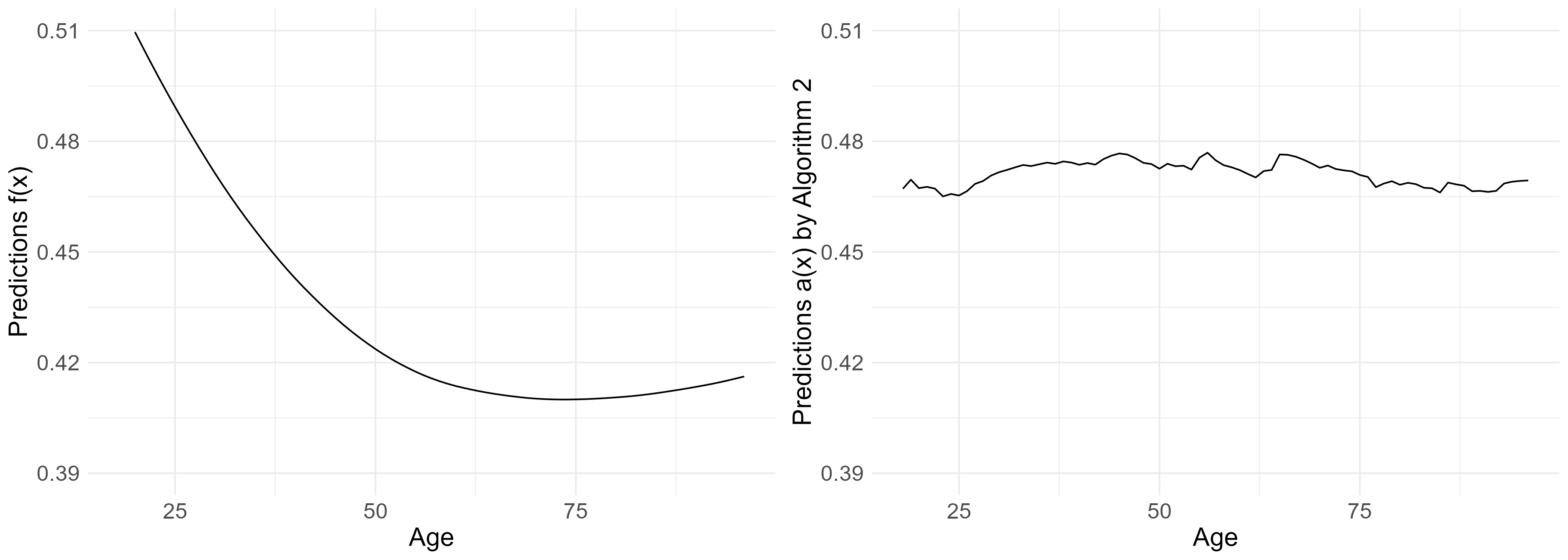}
    \caption{PD Plots for Age before Attack (Left) and after Attack (Right), COMPAS Data}
    \label{fig:pd_age_compas}
\end{figure}

\begin{figure}[h!]
    \centering
    \includegraphics[width=\textwidth]{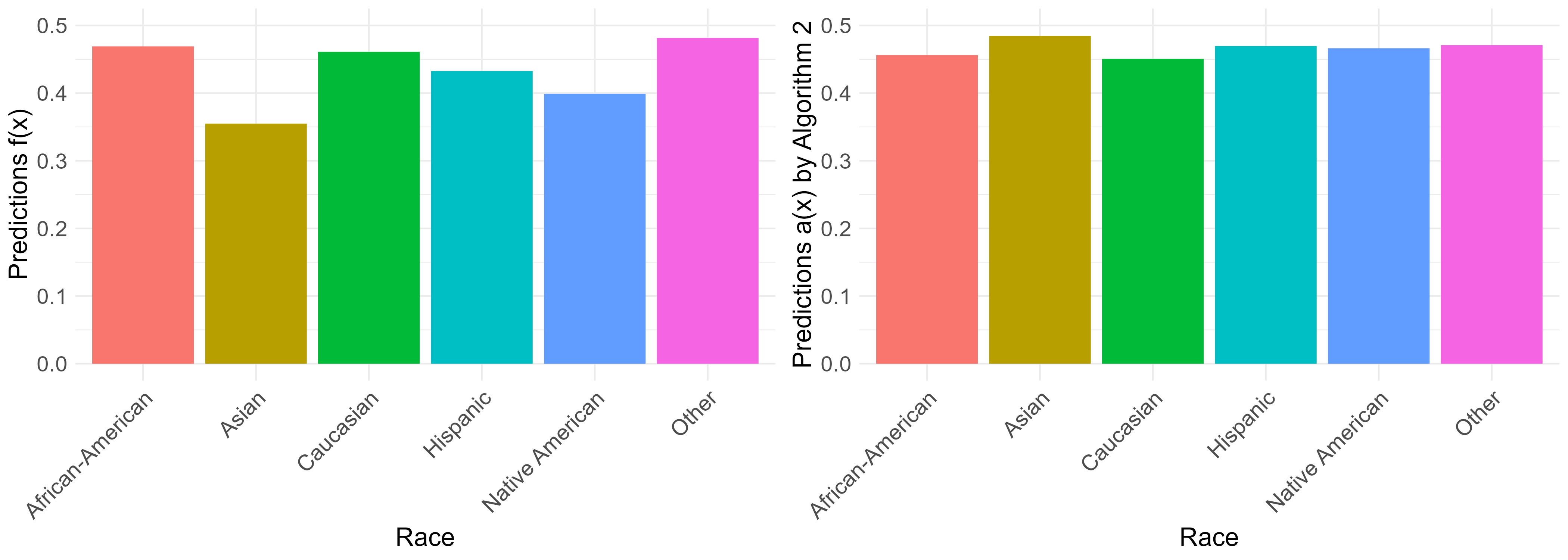}
    \caption{PD Plots for Race before Attack (Left) and after Attack (Right), COMPAS Data}
    \label{fig:pd_race_compas}
\end{figure}

\begin{figure}[h!]
    \centering
    \includegraphics[width=\textwidth]{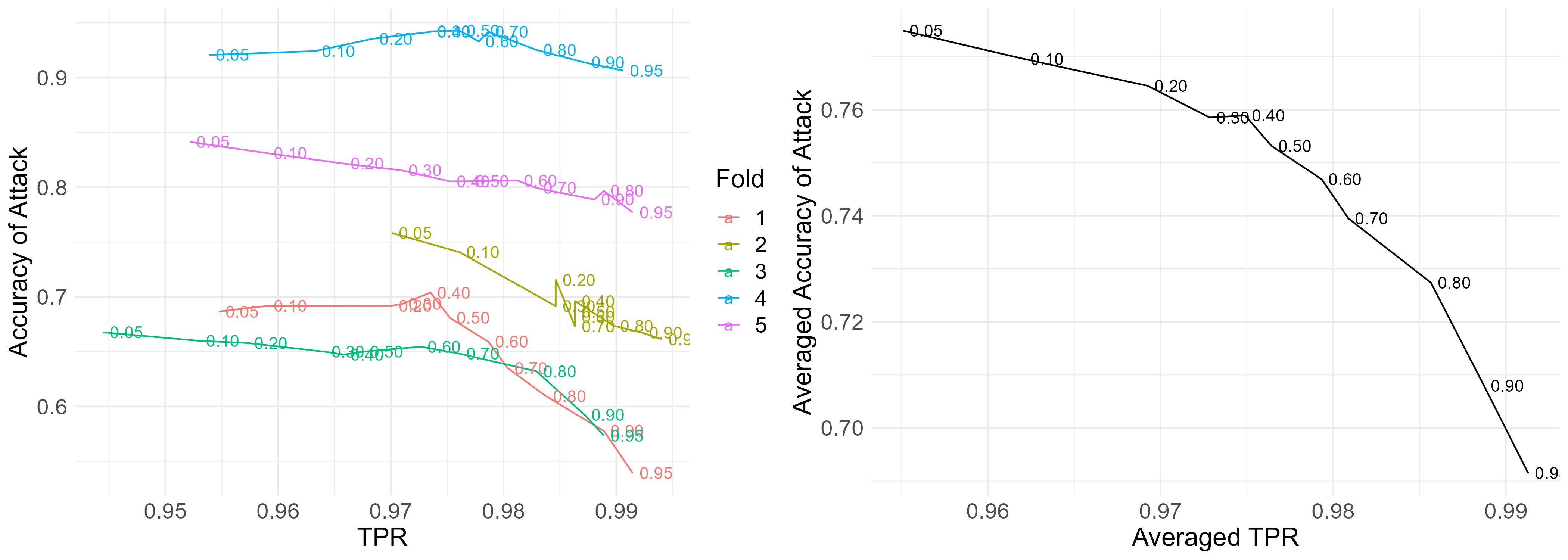}
    \caption{TPR versus Accuracy of Attack across Thresholds \(\{0.05, 0.1, 0.2, \ldots, 0.8, 0.9, 0.95\}\) (labeled next to the lines) for PD Plots of Age, COMPAS Data}
    \label{fig:sd_age_compas}
\end{figure}

\begin{figure}[h!]
    \centering
    \includegraphics[width=\textwidth]{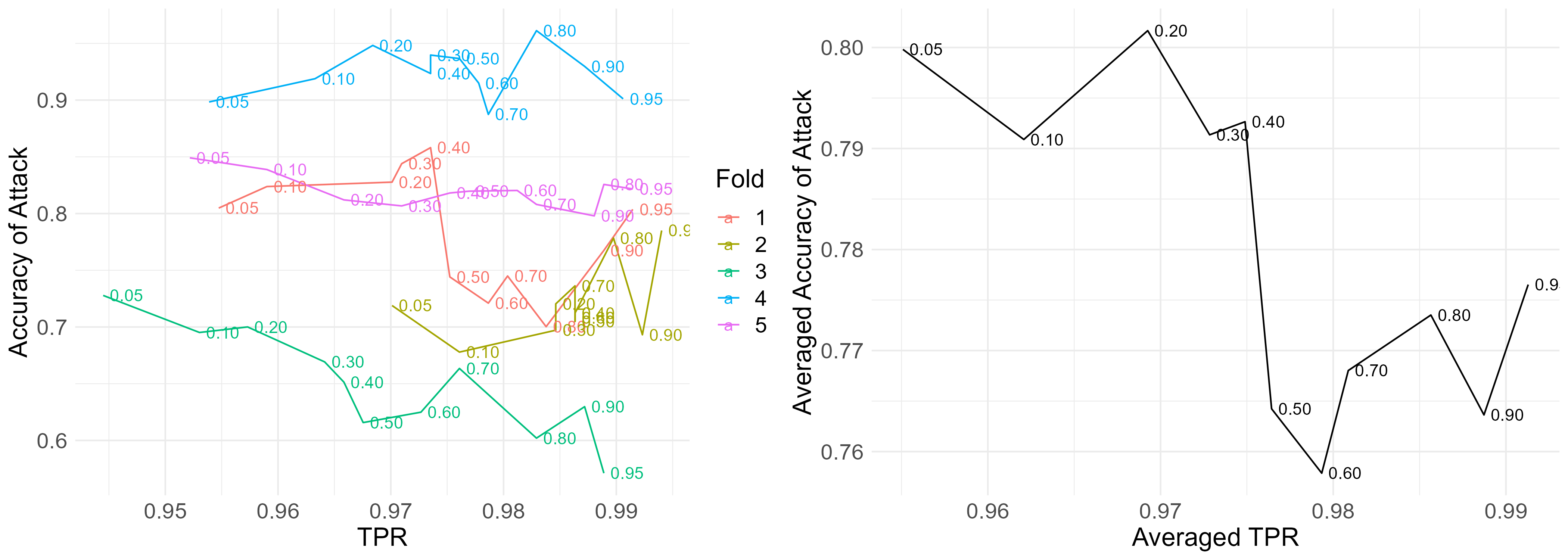}
    \caption{TPR versus Accuracy of Attack across Thresholds \(\{0.05, 0.1, 0.2, \ldots, 0.8, 0.9, 0.95\}\) (labeled next to the lines) for PD Plots of Race, COMPAS Data}
    \label{fig:sd_race_compas}
\end{figure}

We further evaluate our framework using the COMPAS dataset \citep{angwin2016machine}, a popular fairness dataset in fair machine learning literature. This dataset comprises criminal offenders screened in Florida, U.S., during 2013–2014. The response variable, \texttt{two\_year\_recid}, indicates whether a person recidivated within two years after the screening, while sex and race are considered sensitive attributes and the remaining variables serve as predictors. We use the COMPAS data from the fairML R package, which follows the preprocessing steps outlined in \citet{komiyama2018nonconvex}. Variables, summary statistics, and modeling details for the COMPAS dataset used are presented in Appendix \ref{sec:compas_summary} and Appendix \ref{sec:training_nn}.

In our modeling of $f(\textbf{x})$ and $a(\textbf{x})$, we intentionally include sex and race to present an example of manipulating a categorical feature. Our study primarily focuses on age and race as our focal variables, and we use Algorithm \ref{alg:fooling_twopd} to manipulate their PD plots within a unified framework. Similar to the insurance data, we set the target PD outputs as a flat line. Figures \ref{fig:pd_age_compas} and \ref{fig:pd_race_compas} illustrate that older individuals and those of Asian and Native American descent exhibit a lower probability of reoffending. We construct the classifiers $c(\textbf{x})$ and $g(\textbf{x})$ using a process similar to that employed for the insurance data. In the case of the COMPAS data, $f(\textbf{x})$ is trained using a neural network architecture similar to that of $c(\textbf{x})$ and $g(\textbf{x})$. Remarkably, our model maintains strong performance as demonstrated in Figures \ref{fig:sd_age_compas} and \ref{fig:sd_race_compas}, even though the sample size of the COMPAS data (5,836 versus 100,000) is smaller compared to the insurance data and the highest Pearson correlation between the focal variable age and the remaining variables is much smaller (-0.579891 versus 0.920605). For reference, an examination of the correlation matrix for all features used in the auto insurance claims and COMPAS datasets is presented in Appendix \ref{sec:corr_check}.

\subsection{A Simulated Example}
\begin{figure}[h!]
  \centering
  \includegraphics[width=0.45\textwidth]{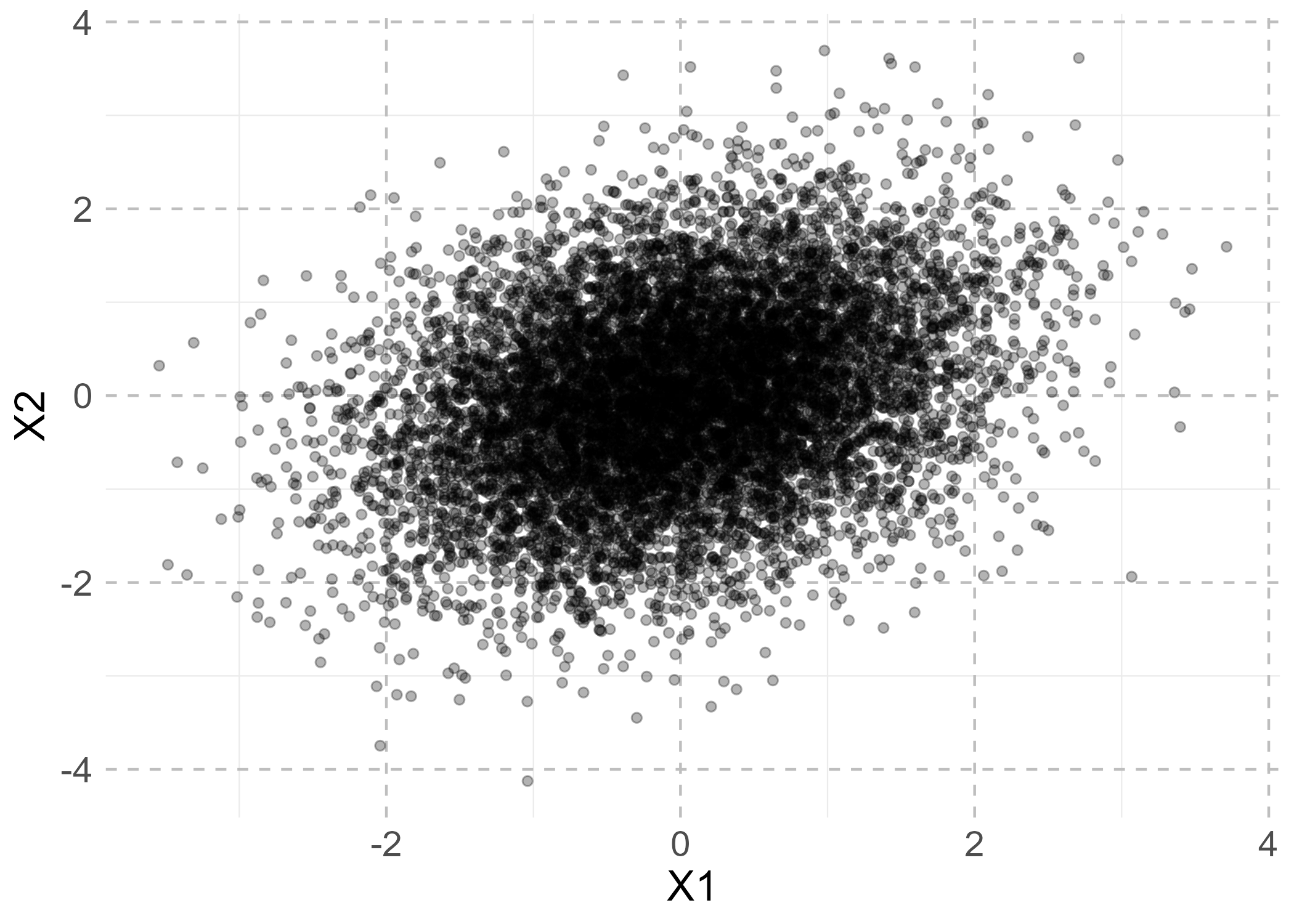}
  \caption{Scatter Plot of Simulated $X_1$ and $X_2$}
  \label{fig:x1_x2}
\end{figure}

As the fooling procedure depends on the interrelationships among features, it is necessary to explore whether only high correlations can facilitate effective fooling. We present a simulated example to demonstrate that an accumulation of weak correlations among features is sufficient for the attacks. The response variable is generated by
\begin{equation}
    y^{(i)} = x^{(i)}_1 + x^{(i)}_2 + x^{(i)}_3 + x^{(i)}_4 + x^{(i)}_5 + 0 \cdot x^{(i)}_6 + \epsilon^{(i)}
\end{equation}
where $\epsilon^{(i)} \sim N(0, 0.5^2)$ adds noise to the process and $x_6$ is set to have no influence on $y$. The features $\{X_j\}^{6}_{j=1}$ are normally distributed with mean $0$ and standard deviation $1$, and a correlation of $0.3$ between each pair. Neural networks were used to construct $c(\textbf{x})$ and $f(\textbf{x})$, but this time, the threshold of $c(\textbf{x})$ is also optimized and fixed at 0.955 after assigning different weights to the two classes during the training of $c(\textbf{x})$, as the training data of $c(\textbf{x})$ is highly imbalanced. Predictions of $c\textbf{(x})$ for correctly classifying real data hold more significance than on uniform data. Figure \ref{fig:x1_x2} shows the scatter plot between the generated $X_1$ and $X_2$, where the correlation of $0.3$ is hardly distinguishable by the eye.

\begin{figure}[h!]
    \centering
    \includegraphics[width=\textwidth]{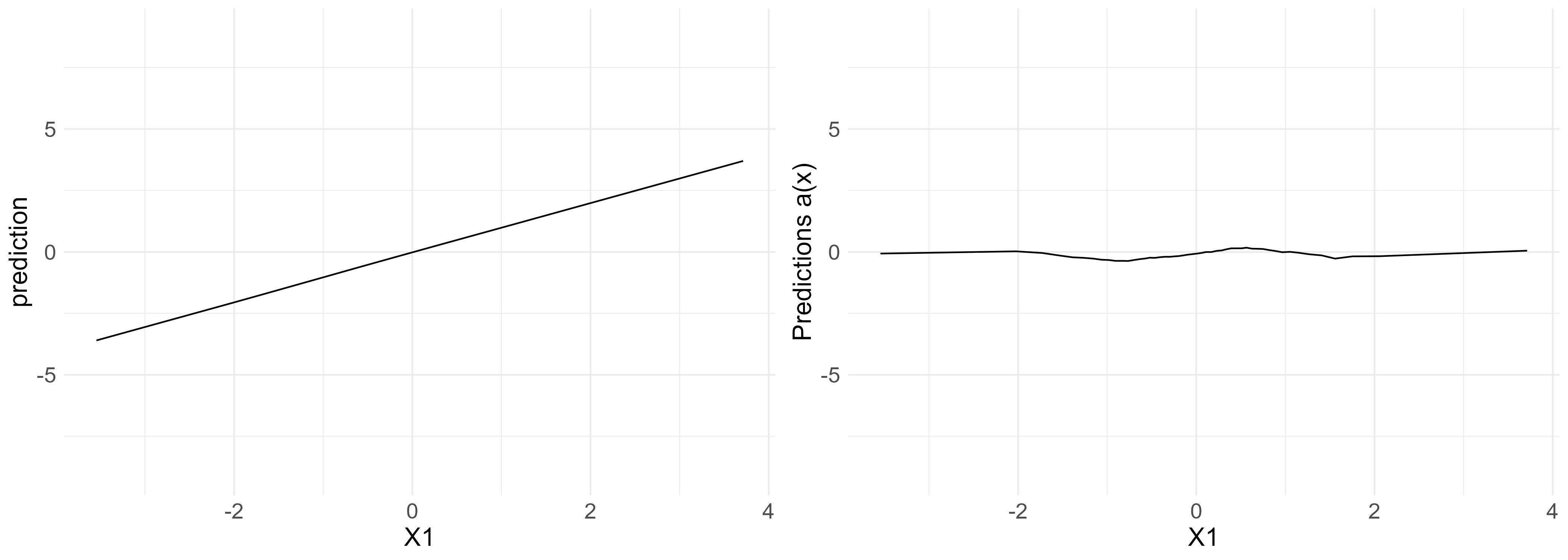}
    \caption{PD Plots for $X_1$ before Attack (Left) and after Attack (Right), Simulated Data}
    \label{fig:pd_x1_simulate}
\end{figure}

\begin{figure}[h!]
    \centering
    \includegraphics[width=\textwidth]{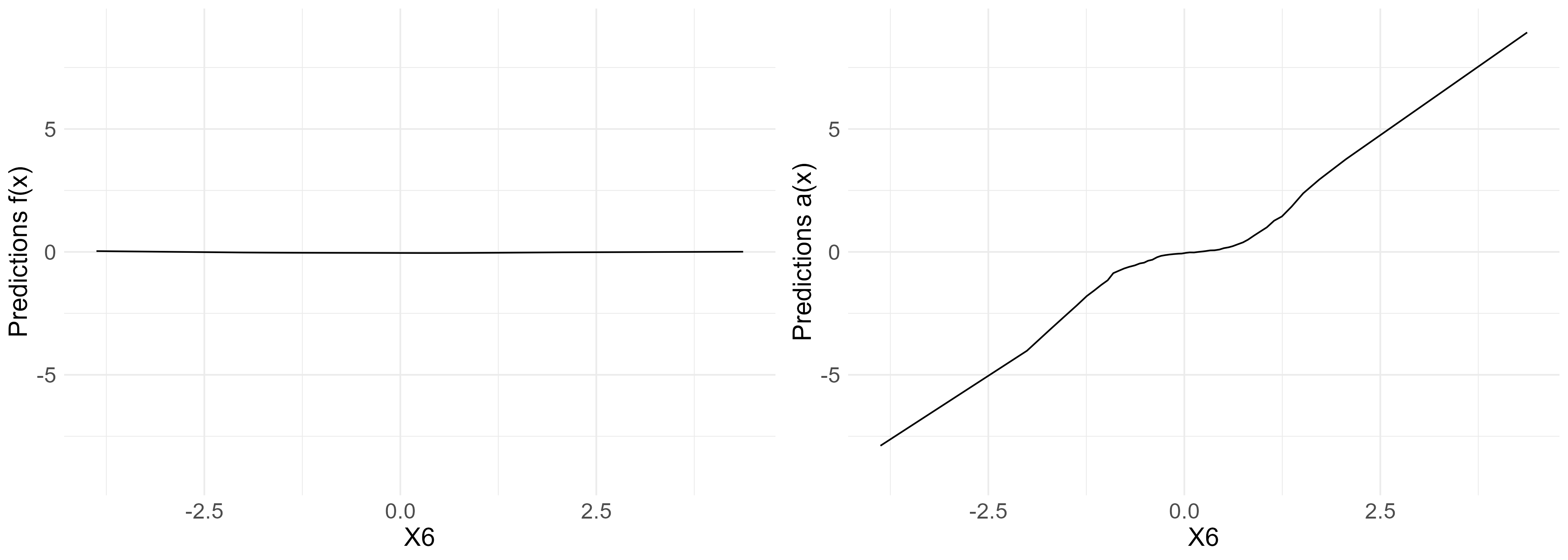}
    \caption{PD Plots for $X_6$ before Attack (Left) and after Attack (Right), Simulated Data}
    \label{fig:pd_x6_simulate}
\end{figure}

We set the target PD outputs as a flat line for $X_1$ and an increasing line with a slope of 2 for $X_6$. The entire process is performed five times on five-fold validation data. The average TPR retained across five folds is $92.09\%$, and the PD plots of features $X_1$ and $X_6$ before and after the attacks are displayed in Figures \ref{fig:pd_x1_simulate} and \ref{fig:pd_x6_simulate}. The observed curvature in the manipulated PD plots can be attributed to the relatively high misclassification errors of $c_1(\textbf{x})$, indicating the challenge in identifying extrapolations for feature values near their mean. Our adversarial framework maintains effectiveness, despite the presence of only weakly correlated features in the data.

\begin{figure}[h!]
    \centering
    \includegraphics[width=\textwidth]{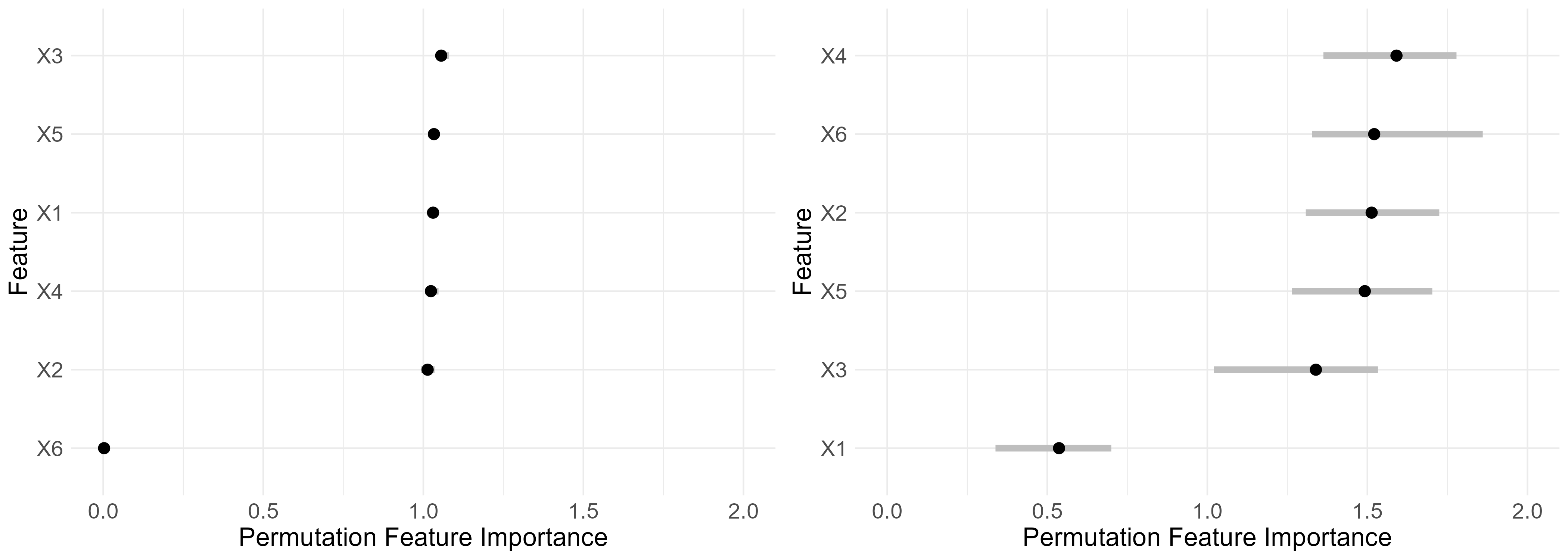}
    \caption{Permutation Feature Importance for $f(\textbf{x})$ before Attack (Left) and $a(\textbf{x})$ after Attack (Right), Simulated Data. The grey lines represent the range from the 10th to the 90th percentile of PFI estimates across fifty runs, with median values indicated by black dots. }
    \label{fig:pfi_simulate}
\end{figure}

In Figure \ref{fig:pfi_simulate}, we explore the permutation feature importance (PFI) of our adversarial algorithm $a(x)$ as a by-product of our adversarial attacks on PD plots. Before the attack, features $X_1$ to $X_5$ demonstrate roughly equal importance in $f(x)$ as we expected. After the attacks targeting $X_1$ and $X_6$, the importance of $X_6$ significantly increases, whereas $X_1$ shows a reduced importance relative to the other features. Additionally, the PFI estimates show increased variability post-attack, attributed to the use of extreme compensating outputs in our fooling algorithm. A similar analysis of PFI for the insurance and COMPAS datasets is presented in Appendix \ref{sec:pfi_other}.

\section{Insights for Regulators and Practitioners}

\textbf{When to use interpretation tools for black-box models?} There are three primary approaches for practitioners to develop interpretable AI models \citep{eiopa2021artificial}: 1) Use traditional interpretable models like Generalized Linear Models (GLMs) or Generalized Additive Models (GAMs); 2) Adopt a hybrid approach by leveraging black-box models solely for feature engineering; 3) Implement black-box models while supplementing them with interpretable tools. We argue that resorting to the third strategy—relying on interpretation methods to explain black-box models—should only be considered if accuracy is critical and the interpretation requirement is relatively low in the application context. Not just PD plots, but all interpretation tools have pitfalls. Practitioners should avoid the overuse of black-box models if interpretable models can achieve the same level of model performance \citep{rudin2019stop}. In certain contexts, alternatives like GAMs can provide similar predictive accuracy as complex black-box models while offering greater interpretability \citep{lou2013accurate, caruana2015intelligible}.

\begin{figure}[h!]
  \centering
  \includegraphics[width=\textwidth]{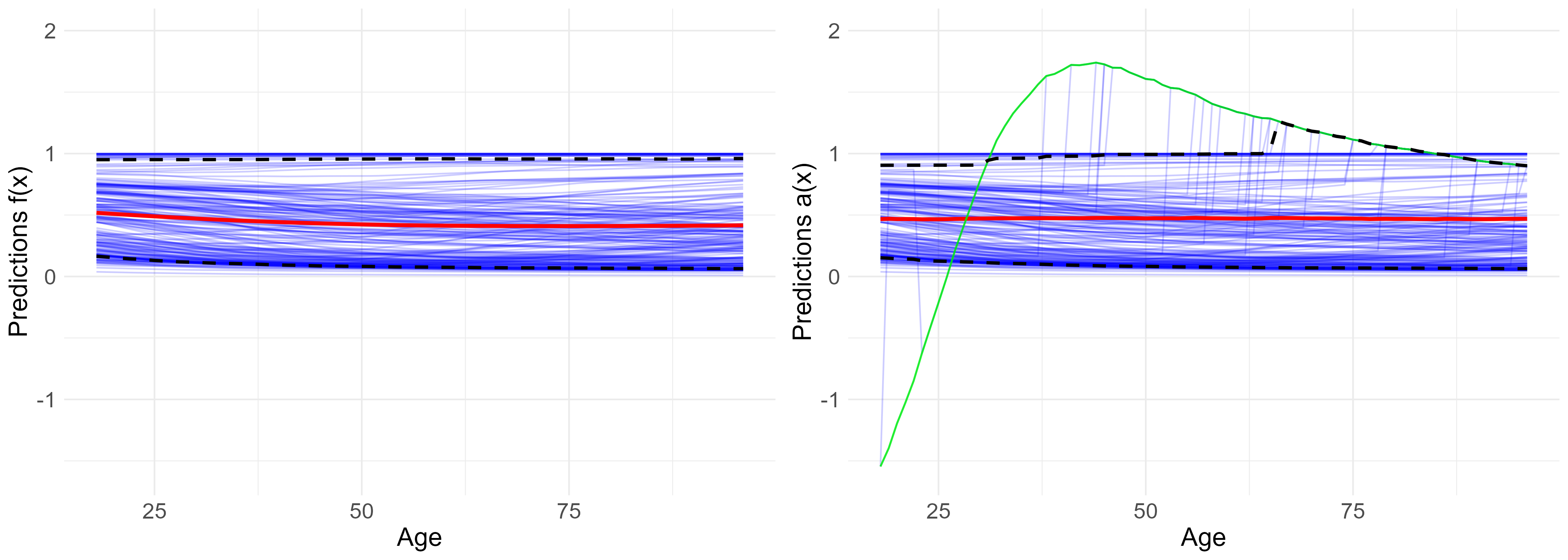}
  \caption{PD Plots (in Red, same as in Figure \ref{fig:pd_age_compas}) for Age with ICE Curves (in Blue) Before (Left) and After (Right) Attack, Including 10th and 90th Percentiles of ICE Curves (in Black) and Compensating Outputs $\hat{\gamma}_{age}$ (in Green), COMPAS Data}
  \label{fig:ice_age}
\end{figure}

\textbf{How to mitigate adversarial attacks on PD plots?} First, enhance PD plots with the addition of Individual Conditional Expectation (ICE) curves \citep{gromping2020model,molnar2022general}, which are first proposed by \citet{goldstein2015peeking}. Mathematically, an ICE curve for a feature $j$ for the $i^{th}$ observation is defined as:
\begin{equation} \label{eqn:ice}
    \mathrm{ICE}^{(i)}_j (x_j) = \hat{f}(x_j, \textbf{x}^{(i)}_{-j})
\end{equation}
Essentially, a PD curve in (\ref{eqn:pdp}) is the average of $n$ ICE curves in (\ref{eqn:ice}) (see Appendix \ref{sec:alternative} for further discussion on ICE plots). As illustrated in Figure \ref{fig:ice_age}, integrating ICE curves with the PD curve for age helps identify anomalies post-attack, although this is effective only when all ICE plots are shown. It is important to note that the 10th and 90th percentile curves do not reveal the attack to nearly the same extent. Additionally, it may also be possible to vary our naive attack to make these plots less visually apparent.

Second, prior to applying interpretation methods, carefully assess the dependencies between features in the data \citep{molnar2022general}. For example, \citet{NAIC2022trees} suggests that regulators obtain a correlation matrix for all predictor variables (see also \citet{SOA2021interpretable}).

Third, to address interpretability concerns and maintain transparency, practitioners may consider using traditional interpretable models like GLMs or GAMs if the benefits of adopting black-box models are not substantial \citep{rudin2019stop}. GAMs can also be used to approximate the $f(\textbf{x})$ in a manner similar to (4.2) and discrepancies between GAMs and their corresponding PDPs may provide reasons to suspect an attack.

Fourth,  practitioners may adopt a hybrid approach that leverages black-box models solely for feature engineering \citep{eiopa2021artificial}.

\textbf{How to use PD plots responsibly in practice?} We advise against employing PD plots as a means to validate the fairness or non-discrimination of sensitive attributes. This is particularly important in adversarial scenarios as characterized by \citet{bordt2022post}, where the stakeholders providing and utilizing interpretation methods have opposing interests and incentives. For instance, while age may be legitimately used, it remains a sensitive variable in black-box modeling, with persistent concerns over its impact on distinct age groups. Our empirical evidence suggests that discriminatory tendencies based on age, especially towards the elderly or youth, can be intentionally hidden, making the model appear neutral through tools like PD plots while preserving nearly all predictions of the original black-box models. Additionally, practitioners should be aware of PD plots' limitations in interpreting black-box models. We advocate for the use of multiple interpretation tools to achieve a more holistic understanding of the model’s feature effects. For a detailed exploration of alternatives to PD plots, we refer readers to Appendix \ref{sec:alternative}.

\section{Conclusions}

\textbf{Potential Impacts of the Work:}
Our framework can be extended to fool permutation-based feature importance. By constructing $g(\textbf{x})$ to reply on some legitimate factors, we can conceal the significant impact of a sensitive variable $X_j$ previously in $f(\textbf{x})$. However, it is difficult to pre-define the target of the desired feature importance value for $X_j$ as we did for PD plots. Permutation-based feature importance can be employed in regulatory fairness checks to assess whether the model assigns significant importance to controversial variables, such as race or sex, in making predictions (a concern analogous to the one presented by \citet{heo2019fooling}
). A real regulatory example of feature importance can be found in California's private passenger automobile insurance regulation (California Code of Regulations, Title 10, Section 2632.5), which stipulates that three mandatory rating variables must carry greater weight in pricing compared to other optional rating variables. Our adversarial framework can potentially hide the significant importance of optional rating variables used and satisfy the regulatory requirements.

\textbf{Limitations of the Work:} First, Our adversarial framework relies on the assumption that the target feature is not independent of other features in the dataset. Although this assumption might not universally hold, it is often plausible in data-rich contexts. As illustrated in our simulated example, our framework remains efficient even with weak feature correlations among features. Second, the success of the attacks is contingent on the level of information available to both attackers and defenders. We make the assumption that attackers (practitioners) have access to all relevant information, while defenders (regulators) have limited information, such as not being able to inspect the underlying codes. Third, there exists a possibility that the compensatory output, denoted by $\gamma_j(x)$, could potentially fall outside the expected domain or range. Fourth, when attempting to fool multiple PD plots, the compensatory output for each feature is set independently before constructing the classifier $c_1(\textbf{x})$. Though it is feasible to calculate these estimates post-training of $c_1(\textbf{x})$ to incorporate its uncertainty estimation, this approach is notably more complex.


\appendix

\section{Datasets Used} \label{sec:data_summary}

\subsection{Data Preparation for Insurance Data} \label{sec:insurance_summary}
As mentioned in the main text, the data preprocessing was conducted in accordance with the methodology proposed by \citet{havrylenko2022detection}. The steps include:
\begin{enumerate}
    \item For the claims dataset \texttt{pg17trainclaim}, calculate the sum of claim numbers grouped by \texttt{id\_client}, \texttt{id\_vehicle}, and \texttt{id\_year} for each row.
    \item Merge the \texttt{pg17trainclaim} dataset with the \texttt{pg17trainpol} dataset using \texttt{id\_client}, \\ \texttt{id\_vehicle}, and \texttt{id\_year} as keys.
    \item Replace NA values in \texttt{claim\_nb} with zero.
    \item Filter out observations where the age of the driving license (\texttt{drv\_age\_lic1}) is less than the driver's age (\texttt{drv\_age1}) minus 17.
    \item Remove observations with NA values in \texttt{vh\_age}.
    \item Impute mean values for zero entries in \texttt{vh\_value} and \texttt{vh\_weight}, based on \texttt{vh\_make} and \texttt{vh\_model} categories.
    \item Reduce the 101 categories of \texttt{vh\_make} to 18 by merging those with similar average responses, applying k-means clustering for the aggregation. Label this newly aggregated feature as \texttt{vh\_makenew}.
\end{enumerate}

Following these preprocessing steps, the merged dataset comprises 99,918 observations. Due to the absence of exposure information in the dataset, we introduce the feature \texttt{annual\_exposure} and assume an equal exposure value of 1 for all observations. Below is a list of all features available in the insurance dataset:

\begin{lstlisting}
'data.frame':	99918 obs. of  31 variables:
 $ pol_bonus       : num  0.5 0.5 0.5 0.5 0.5 0.5 0.5 0.5 0.64 0.5 ...
 $ pol_coverage    : Factor w/ 4 levels "Maxi","Median1",..: 1 1 1 3 1 2 1 1 3 1 ...
 $ pol_duration    : int  29 3 2 22 16 5 5 2 5 26 ...
 $ pol_sit_duration: int  9 1 2 1 4 1 3 2 1 6 ...
 $ pol_pay_freq    : Factor w/ 4 levels "Biannual","Monthly",..: 1 1 4 4 1 2 1 1 2 1 ...
 $ pol_payd        : Factor w/ 2 levels "No","Yes": 1 1 1 1 1 1 1 1 1 1 ...
 $ pol_usage       : Factor w/ 4 levels "AllTrips","Professional",..: 3 3 4 4 3 4 3 3 4 4 ...
 $ pol_insee_code  : Factor w/ 17794 levels "01001","01004",..: 6424 17569 17550 15522 6911 14819 6914 6498 16274 14572 ...
 $ drv_drv2        : Factor w/ 2 levels "No","Yes": 1 1 1 2 2 1 1 1 1 2 ...
 $ drv_age1        : int  85 69 37 81 62 68 77 64 38 59 ...
 $ drv_age2        : int  0 0 0 21 68 0 0 0 0 33 ...
 $ drv_sex1        : Factor w/ 2 levels "F","M": 2 2 2 2 1 2 2 2 2 2 ...
 $ drv_sex2        : Factor w/ 3 levels "","F","M": 1 1 1 2 3 1 1 1 1 2 ...
 $ drv_age_lic1    : int  62 39 18 54 37 40 55 37 19 41 ...
 $ drv_age_lic2    : int  0 0 0 3 48 0 0 0 0 15 ...
 $ vh_age          : int  10 4 11 16 11 14 7 11 9 6 ...
 $ vh_cyl          : int  1587 2149 1991 1781 1598 1769 1870 1595 1997 1997 ...
 $ vh_din          : int  98 170 150 90 108 60 108 101 109 90 ...
 $ vh_fuel         : Factor w/ 3 levels "Diesel","Gasoline",..: 2 1 2 2 2 1 1 2 1 1 ...
 $ vh_make         : Factor w/ 101 levels "ACL","ALFA ROMEO",..: 71 60 16 98 76 71 76 7 71 71 ...
 $ vh_model        : Factor w/ 1023 levels "+4","10","100",..: 75 311 1014 514 599 40 599 214 76 725 ...
 $ vh_sale_begin   : int  10 4 12 18 13 28 10 16 9 9 ...
 $ vh_sale_end     : int  9 2 11 15 11 18 6 13 7 7 ...
 $ vh_speed        : int  182 229 210 180 195 155 193 191 183 163 ...
 $ vh_type         : Factor w/ 2 levels "Commercial","Tourism": 2 2 2 2 2 2 2 2 2 2 ...
 $ vh_value        : num  20700 34250 28661 14407 16770 ...
 $ vh_weight       : num  1210 1510 1270 1020 1230 ...
 $ claim_nb        : num  0 0 0 0 0 0 0 0 1 0 ...
 $ claim_amount    : num  NA NA NA NA NA ...
 $ vh_makenew      : Factor w/ 18 levels "Group 1","Group 10",..: 2 7 15 15 1 2 1 7 2 2 ...
 $ annual_exposure : num  1 1 1 1 1 1 1 1 1 1 ...
\end{lstlisting}

Please note that the accuracy of $f(x)$ does not directly impact the effectiveness of our fooling algorithm. For demonstration purposes, we construct a claim frequency model, denoted as $f(x)$, using the insurance dataset. Besides \texttt{annual\_exposure} and \texttt{claim\_nb} as target or offset variables, we have selected 14 features for the model. These include four policy characteristics (\texttt{pol\_bonus}, \texttt{pol\_coverage}, \texttt{pol\_duration}, \texttt{pol\_sit\_duration}), three primary policyholder characteristics (\texttt{drv\_age1}, \texttt{drv\_sex1}, \texttt{drv\_age\_lic1}), and seven vehicle characteristics (\texttt{vh\_age}, \texttt{vh\_cyl}, \texttt{vh\_din}, \texttt{vh\_fuel}, \texttt{vh\_speed}, \texttt{vh\_value}, \texttt{vh\_makenew}). Table \ref{tab:variables_pg17} lists the variables used in the insurance dataset, with descriptions extracted from \citet{dutang2019casdatasets}.

\begin{table}[h]
\centering
\begin{tabular}{lp{0.6\textwidth}} 
\toprule
\textbf{Variable Name} & \textbf{Description} \\
\midrule
\textbf{Response Variables} &  \\
  \texttt{claim\_nb} & The claim number.\\
\addlinespace
\textbf{Explanatory Variables} &  \\
  \texttt{pol\_bonus} & The policy bonus (French no-claim discount). \\
  \texttt{pol\_coverage} & The coverage category, including 4 types : Mini, Median1, Median2 and Maxi, in this order.\\
  \texttt{pol\_duration} & The policy duration. \\
  \texttt{pol\_sit\_duration} & The policy current endorsement (situation) duration.\\
\addlinespace
  \texttt{drv\_age1} & The driver age of the 1st driver. This is a focal variable for adversarial attacks.\\
  \texttt{drv\_sex1} & The driver sex of the 1st driver.\\
  \texttt{drv\_age\_lic1} & The age of the driving license of the 1st driver.\\
\addlinespace
  \texttt{vh\_age} & The vehicle age. \\
  \texttt{vh\_cyl} & The engine cylinder displacement. \\
  \texttt{vh\_din} & A representation of the motor power. \\
  \texttt{vh\_fuel} & The vehicle fuel type. \\
  \texttt{vh\_speed} & The vehicle maximum speed (km/h), as stated by the manufacturer.\\
  \texttt{vh\_value} & The vehicle’s value (replacement value). This is a focal variable for adversarial attacks. \\
  \texttt{vh\_makenew} & The vehicle carmaker, which are reduced from 101 categories to 18 categories by merging those with similar average responses.\\
\bottomrule
\end{tabular}
\caption{Variables Used in $f(\textbf{x})$ for the Insurance Data}
\label{tab:variables_pg17}
\end{table}

\subsection{COMPAS Data} \label{sec:compas_summary}

As mentioned in the main text, we directly use the COMPAS data from the fairML R package, which follows the preprocessing steps outlined in \citet{komiyama2018nonconvex}. The dataset comprises 5,855 observations. Below is a list of all features available in the COMPAS dataset:

\begin{lstlisting}
'data.frame':	5855 obs. of  16 variables:
 $ age            : num  69 34 24 44 41 39 21 27 23 37 ...
 $ juv_fel_count  : num  0 0 0 0 0 0 0 0 0 0 ...
 $ decile_score   : num  1 3 4 1 6 1 3 4 6 1 ...
 $ juv_misd_count : num  0 0 0 0 0 0 0 0 0 0 ...
 $ juv_other_count: num  0 0 1 0 0 0 0 0 0 0 ...
 $ v_decile_score : num  1 1 3 1 2 1 5 4 4 1 ...
 $ priors_count   : num  0 0 4 0 14 0 1 0 3 0 ...
 $ sex            : Factor w/ 2 levels "Female","Male": 2 2 2 2 2 1 2 2 2 1 ...
 $ two_year_recid : Factor w/ 2 levels "No","Yes": 1 2 2 1 2 1 2 1 2 1 ...
 $ race           : Factor w/ 6 levels "African-American",..: 6 1 1 6 3 3 3 3 1 3 ...
 $ c_jail_in      : num  0.801 0.784 0.791 0.811 0.818 ...
 $ c_jail_out     : num  0.801 0.785 0.791 0.811 0.818 ...
 $ c_offense_date : num  0.801 0.784 0.791 0.811 0.818 ...
 $ screening_date : num  0.801 0.785 0.791 0.811 0.818 ...
 $ in_custody     : num  0.829 0.784 0.796 0.811 0.821 ...
 $ out_custody    : num  0.83 0.785 0.796 0.811 0.823 ...
\end{lstlisting}

In addition, descriptions for variables in the COMPAS dataset are provided in Table 2, extracted from the fairML R package.

\begin{table}[h]
\centering
\begin{tabular}{lp{0.6\textwidth}} 
\toprule
\textbf{Variable Name} & \textbf{Description} \\
\midrule
\textbf{Response Variable} &  \\
  \texttt{two\_year\_recid} & A factor with two levels "Yes" and "No" (if the person has recidivated within two years). \\
\addlinespace
\textbf{Explanatory Variables} &  \\
  \texttt{sex} & A factor with levels "Female" and "Male". This is a focal variable for adversarial attacks.\\
  \texttt{race} & A factor encoding the race of the person. This is a focal variable for adversarial attacks.\\
  \texttt{juv\_fel\_count} & The number of juvenile felonies. \\
  \texttt{decile\_score} & The decile of the COMPAS score. \\
  \texttt{juv\_misd\_count} & The number of juvenile misdemeanors. \\
  \texttt{juv\_other\_count} & The number of prior juvenile convictions that are not considered either felonies or misdemeanors. \\
  \texttt{v\_decile\_score} & The predicted decile of the COMPAS score. \\
  \texttt{priors\_count} & The number of prior crimes committed. \\
  \texttt{c\_jail\_in} & The date in which the person entered jail (normalized between 0 and 1). \\
  \texttt{c\_jail\_out} & The date in which the person was released from jail (normalized between 0 and 1). \\
  \texttt{c\_offense\_date} & The date the offense was committed. \\
  \texttt{screening\_date} & The date in which the person was screened (normalized between 0 and 1). \\
  \texttt{in\_custody} & The date in which the person was placed in custody (normalized between 0 and 1). \\
  \texttt{out\_custody} & The date in which the person was released from custody (normalized between 0 and 1). \\
\bottomrule
\end{tabular}
\caption{Variables Used in $f(\textbf{x})$ for the COMPAS Data}
\label{tab:variables_compas}
\end{table}

\section{Correlations of Features within Datasets Used} \label{sec:corr_check}

Our adversarial framework relies on the assumption that the target feature is not independent of other features in the dataset. In this section, we use Spearman's rank correlation to present and examine the correlation matrix for all features used in the auto insurance claims and COMPAS datasets. For nominal variables, we approximate correlations by transforming the data into ordinal form, reordering categories based on their respective risk levels. This transformation applies to \texttt{pol\_coverage}, \texttt{vh\_fuel}, and \texttt{vh\_makenew} in the insurance data, and \texttt{race} in the COMPAS data.

\begin{figure}[h!]
    \centering
    \includegraphics[width=\textwidth]{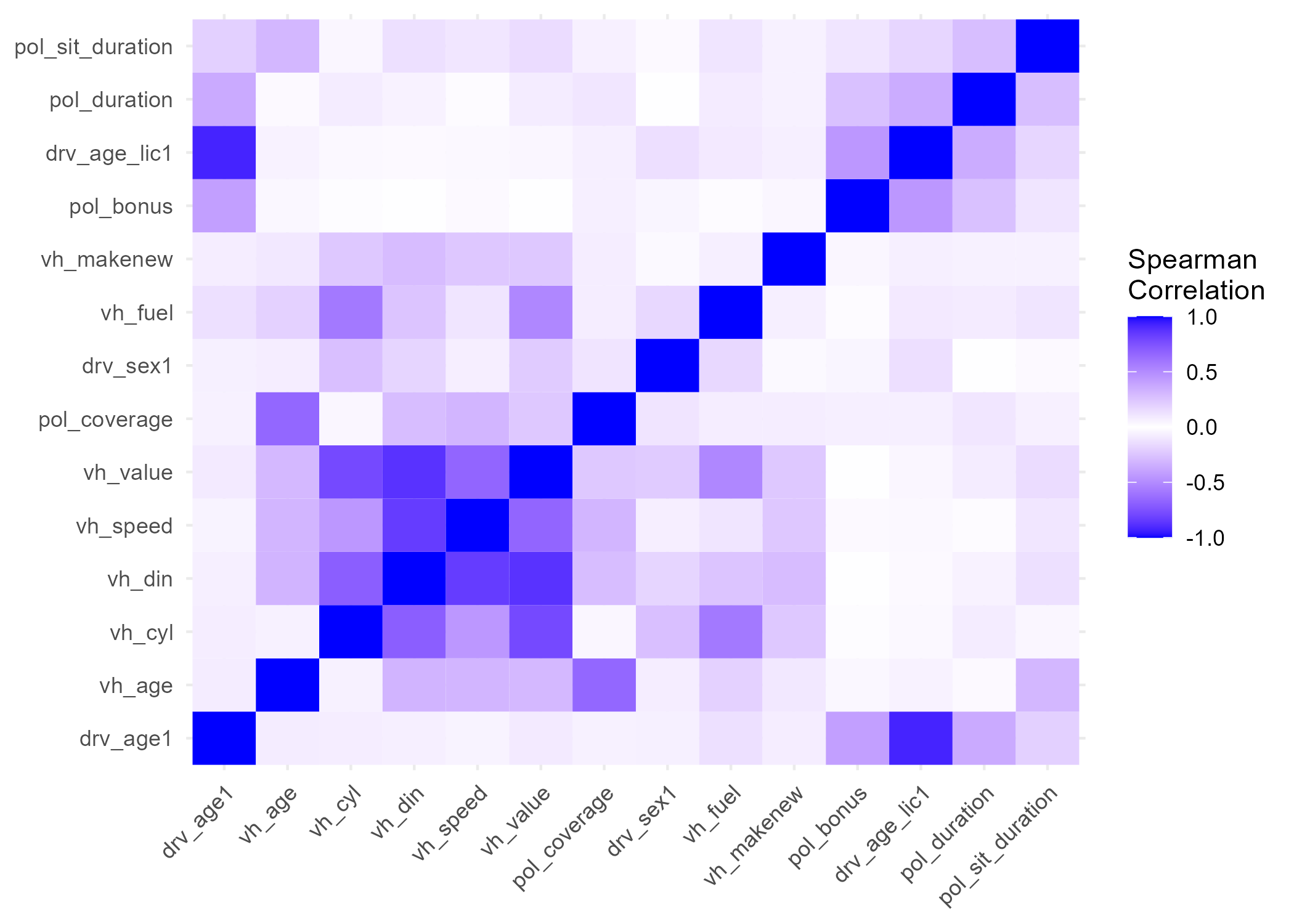}
    \caption{Correlation Matrix for the Insurance Data}
    \label{fig:corr_pg17}
\end{figure}

As depicted in Figure \ref{fig:corr_pg17} for the insurance data, the driver's age (\texttt{drv\_age1}) shows the highest correlation with the age of the driving license (\texttt{drv\_age\_lic1}), followed by the policy bonus (\texttt{pol\_bonus}) and the policy duration (\texttt{pol\_duration}). Furthermore, the vehicle characteristic variables exhibit mutual correlations.

\newpage

\begin{figure}[h!]
    \centering
    \includegraphics[width=\textwidth]{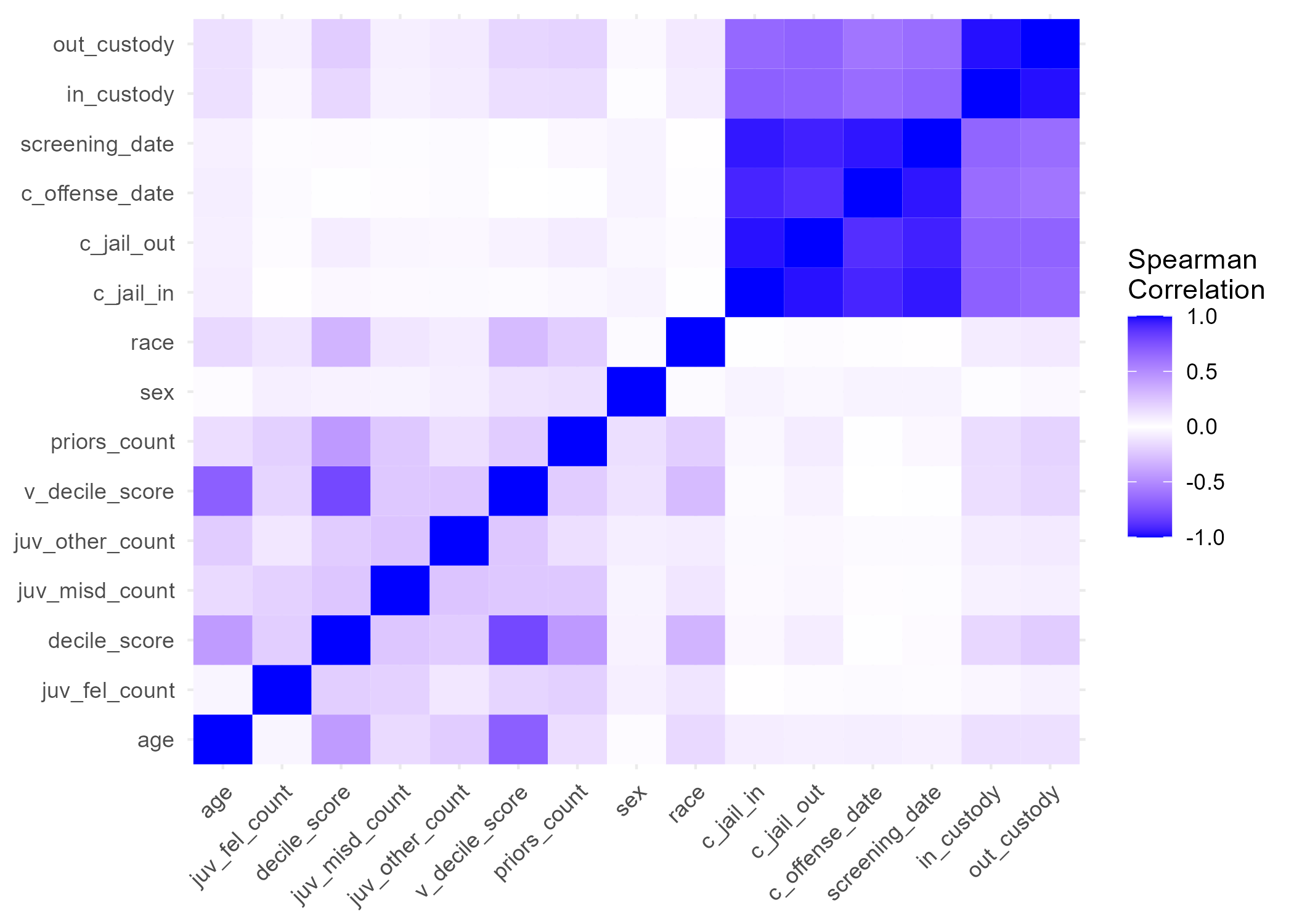}
    \caption{Correlation Matrix for COMPAS Data}
    \label{fig:corr_compas}
\end{figure}

For the COMPAS data, as depicted in Figure \ref{fig:corr_compas}, it is expected that some date-related features exhibit high correlations amongst themselves (\texttt{c\_jail\_in}, \texttt{c\_jail\_out}, \texttt{c\_offense\_date}, \texttt{screening\_date}, \texttt{in\_custody}, \texttt{out\_custody}). The variable \texttt{v\_decile\_score}, a continuous variable containing the predicted decile of the COMPAS score, shows the most significant correlation with age.

\newpage

\section{More Implementation Details of Adversarial Framework $a(\textbf{x})$} \label{sec:more_cx}

In this work, our training data is split into five-folds, and each time, four of the folds are combined as the training data, and the remaining fold is the testing data.

\subsection{The Training of $c(\textbf{x})$}

The training process of $c(\textbf{x})$ involves generating an augmenting sample $\tilde{X}$, which, as outlined in the main text, is drawn randomly and independently among features according to the uniform $\mathbb{P}(\tilde{\textbf{x}})$. The number of observations in $\tilde{X}$ is 30 times the sample size of the insurance data, and 100 times for both the COMPAS and simulated data. Specifically, for each row in $\tilde{X}$:

\begin{itemize}
    \item For categorical variables, each feature value $\tilde{x}^{(i)}_j$ is randomly generated from all possible categories of the respective feature.
    \item For numerical features, each feature value $\tilde{x}^{(i)}_j$ is randomly generated from all unique values observed within the feature in the training data (for instance, the vehicle value in the insurance data).
\end{itemize}

This approach to generating numerical features is specifically designed to avoid the issue of extrapolation within the range of the training data, ensuring that any unobserved values not present in $X$ are excluded. Consequently, $c(\textbf{x})$ is constructed to avoid relying on the unique value of a feature for distinguishing between uniform and real data.

\subsection{The Selection of Grid Values for PD Plots} \label{sec:select_grid}

For the continuous feature (\texttt{vh\_value} in the insurance data), the grid values are selected based on quantiles of the feature distribution, instead of using equally spaced points. This approach ensures that the grid points align with the data's distribution. For categorical or discrete features, we employ all available categories or unique values to construct the PD plot.

\section{Training of $f(\textbf{x})$, $c(\textbf{x})$ and $g(\textbf{x})$ Models} \label{sec:training_models}

\subsection{Training of XGBoost Models} \label{sec:training_xgboost}

\begin{table}[ht]
\centering
\caption{XGBoost Hyperparameter Optimization}
\label{tab:xgboost_hyperparams}
\begin{tabular}{lll}
\hline
Hyperparameter       & Values          & Optimal Value \\ \hline
max\_depth           & 2, 3, 4, 5      & 3   \\
eta                  & 0.01, 0.05, 0.1 & 0.01   \\
gamma                & 0, 0.1, 0.5     & 0   \\
subsample            & 0.6, 0.8, 1     & 0.8   \\
colsample\_bytree    & 0.6, 0.8, 1     & 0.8   \\
min\_child\_weight   & 1, 3, 5         & 1   \\ \hline
\end{tabular}
\end{table}

For the claim count model \(f(\mathbf{x})\) applied to insurance data, we fit the XGBoost model, using the \texttt{xgboost} package \citep{chen2022xgboost} in R. Hyperparameter tuning was performed via grid search, with the optimal parameters detailed in Table \ref{tab:xgboost_hyperparams}. It is important to note that the accuracy of \(f(\mathbf{x})\) does not directly impact the effectiveness of our adversarial framework \(a(\mathbf{x})\).

\subsection{Training of Neural Network Models} \label{sec:training_nn}
\begin{table}[htbp]
\centering
\small
\caption{Summary of Neural Network Architectures}
\label{tab:nn_architecture}
\begin{tabular}{>{\raggedright}p{1.5cm} >{\raggedright}p{1cm} >{\raggedright}p{3.2cm} > {\raggedright}p{4.2cm} > {\raggedright\arraybackslash}p{4cm}}
\toprule
\textbf{Dataset} & \textbf{Model} & \textbf{Structure (Nodes per Layer)} & \textbf{Techniques} & \textbf{Activation Functions} \\
\midrule
Insurance & $c(\textbf{x})$ & [40, 20, 10, 1] & Batch Norm, Dropout (0.2) & ReLU (Output: Sigmoid) \\
Insurance & $g(\textbf{x})$ & [40, 20, 10, 3] & Batch Norm, Dropout (0.2) & ReLU (Output: Softmax) \\
COMPAS & $c(\textbf{x})$ & [40, 10, 1] & Batch Norm, Dropout (0.2) & ReLU (Output: Sigmoid) \\
COMPAS & $g(\textbf{x})$ & [40, 10, 3] & Batch Norm, Dropout (0.2) & ReLU (Output: Softmax) \\
COMPAS & $f(\textbf{x})$ & [40, 10, 1] & Batch Norm, Dropout (0.2) & ReLU (Output: Sigmoid) \\
Simulated & $c(\textbf{x})$ & [20, 10, 1] & Batch Norm, Dropout (0.2) & ReLU (Output: Sigmoid) \\
Simulated & $g(\textbf{x})$ & [20, 10, 3] & Batch Norm, Dropout (0.2) & ReLU (Output: Softmax) \\
Simulated & $f(\textbf{x})$ & [20, 10, 1] &  None & ReLU (Output: Linear) \\
\bottomrule
\end{tabular}
\end{table}

The neural network models were trained using the \texttt{keras} and \texttt{tensorflow} packages in R. Training of each neural network model starts with a smaller, commonly used model architecture, gradually incorporating additional features and enhancements. For further tuning strategies, we direct interested readers to \citet{tuningplaybookgithub}. A summary of the neural network models' architecture is presented in Table \ref{tab:nn_architecture}. Moreover, techniques such as early stopping and learning rate scheduling are employed to mitigate overfitting and enhance model performance.

\section{Other Experimental Results}

\subsection{Experimental Results Using Algorithm 1} \label{sec:using_alg1}

\begin{figure}[h!]
    \centering
    \includegraphics[width=\textwidth]{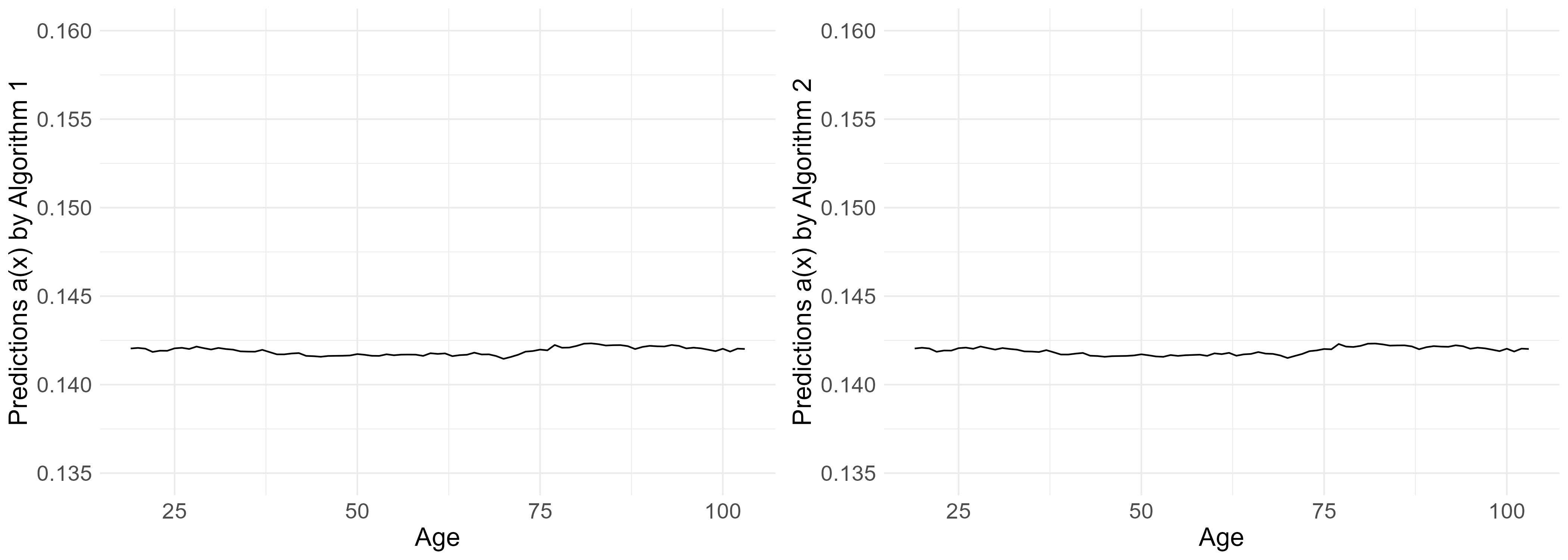}
    \caption{PD Plots for Age after Attack Using Algorithm 1 (Left) and Algorithm 2 (Right), pg17 Data}
    \label{fig:pd_agealg12_pg17}
\end{figure}

\begin{figure}[h!]
    \centering
    \includegraphics[width=\textwidth]{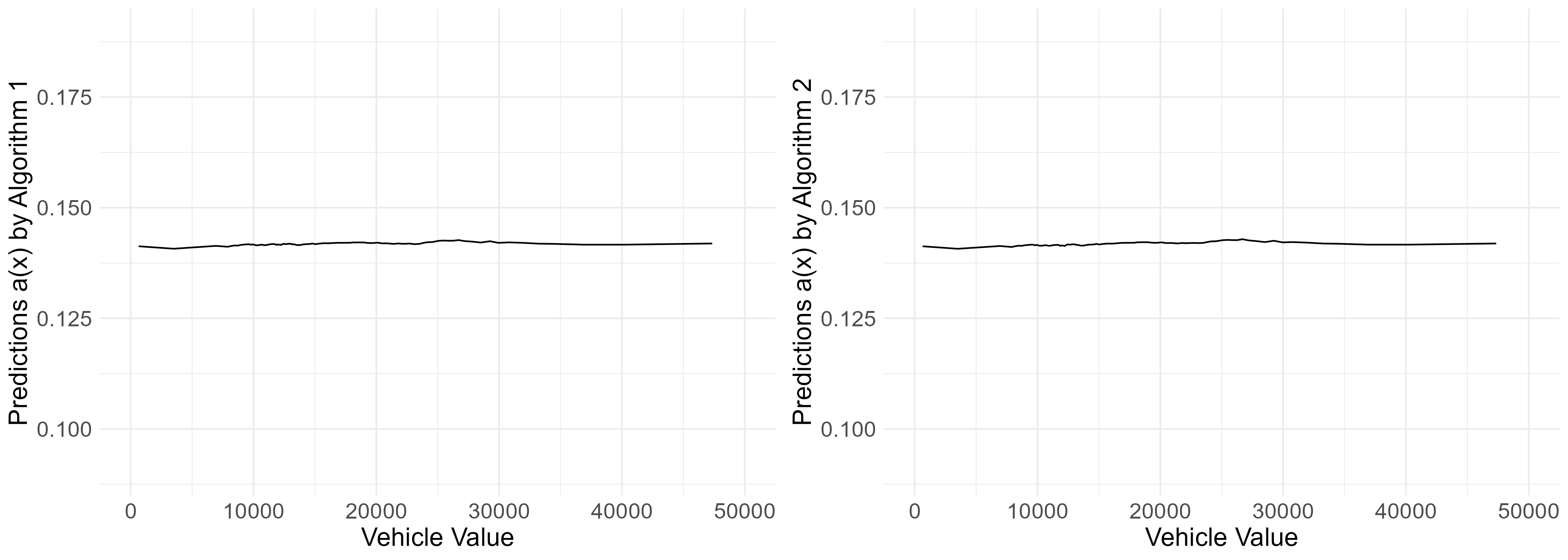}
    \caption{PD Plots for Vehicle Value after Attack Using Algorithm 1 (Left) and Algorithm 2 (Right), pg17 Data}
    \label{fig:pd_vhvalg12_pg17}
\end{figure}

\begin{figure}[h!]
    \centering
    \includegraphics[width=\textwidth]{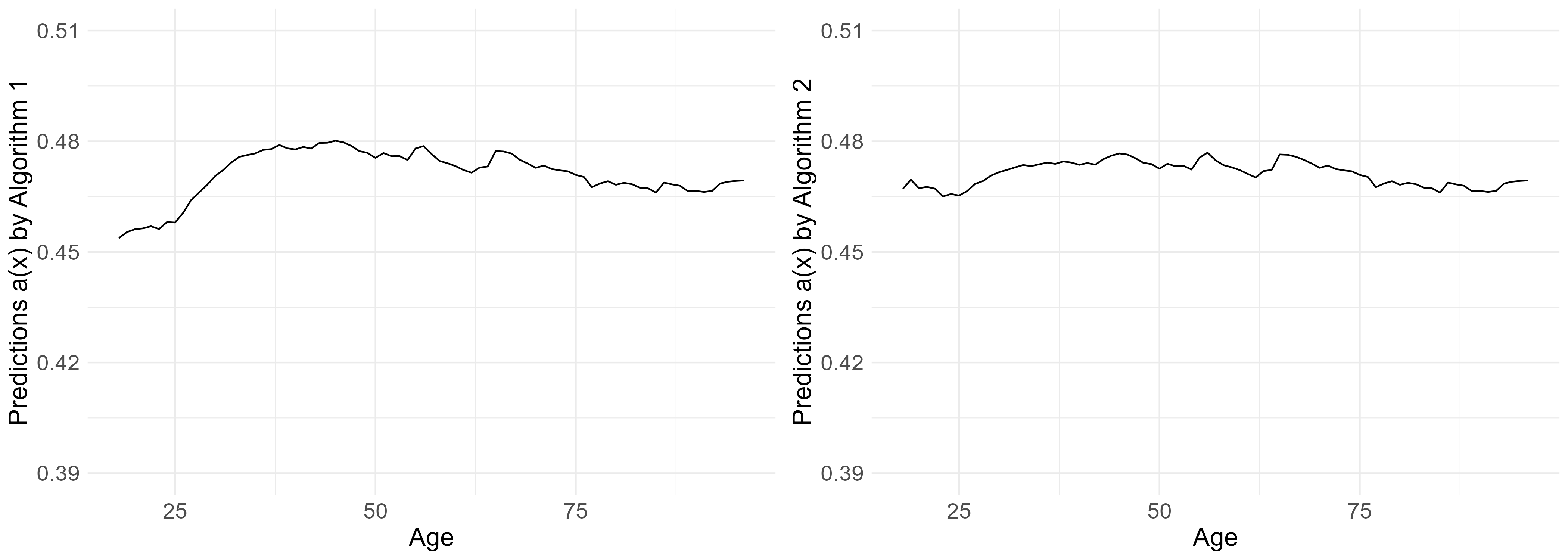}
    \caption{PD Plots for Age after Attack Using Algorithm 1 (Left) and Algorithm 2 (Right), COMPAS Data}
    \label{fig:pd_agealg12_compas}
\end{figure}

\begin{figure}[h!]
    \centering
    \includegraphics[width=\textwidth]{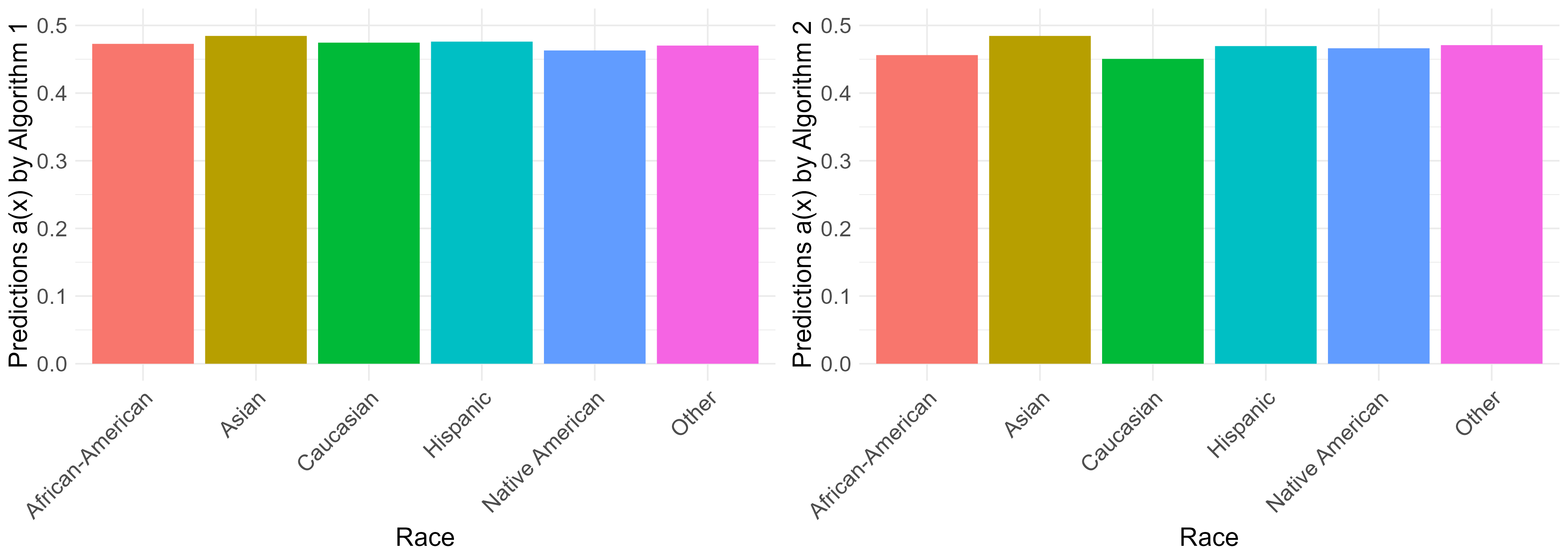}
    \caption{PD Plots for Race after Attack Using Algorithm 1 (Left) and Algorithm 2 (Right), COMPAS Data}
    \label{fig:pd_racealg12_compas}
\end{figure}

In this appendix, we present the experimental results for the insurance and COMPAS datasets using Algorithm 1 in Figures \cref{fig:pd_agealg12_pg17,fig:pd_vhvalg12_pg17,fig:pd_agealg12_compas,fig:pd_racealg12_compas}. Generally, Algorithm 2 does not show significantly poorer performance in terms of stability for the chosen fold and threshold (i.e., holding out fold 1 and setting the threshold of $c(\textbf{x})$ to 0.5). The marginally better performance of Algorithm 2 on the COMPAS data is likely due to the errors in $\hat{g}_c(\textbf{x})$ accidentally compensating for the errors in estimating $\hat{\rho}_j$, $\hat{\lambda}_j$ and $\hat{\gamma}_j$.

\subsection{Permutation Feature Importance (PFI) for Insurance and COMPAS Data} \label{sec:pfi_other}

\begin{figure}[h!]
    \centering
    \includegraphics[width=\textwidth]{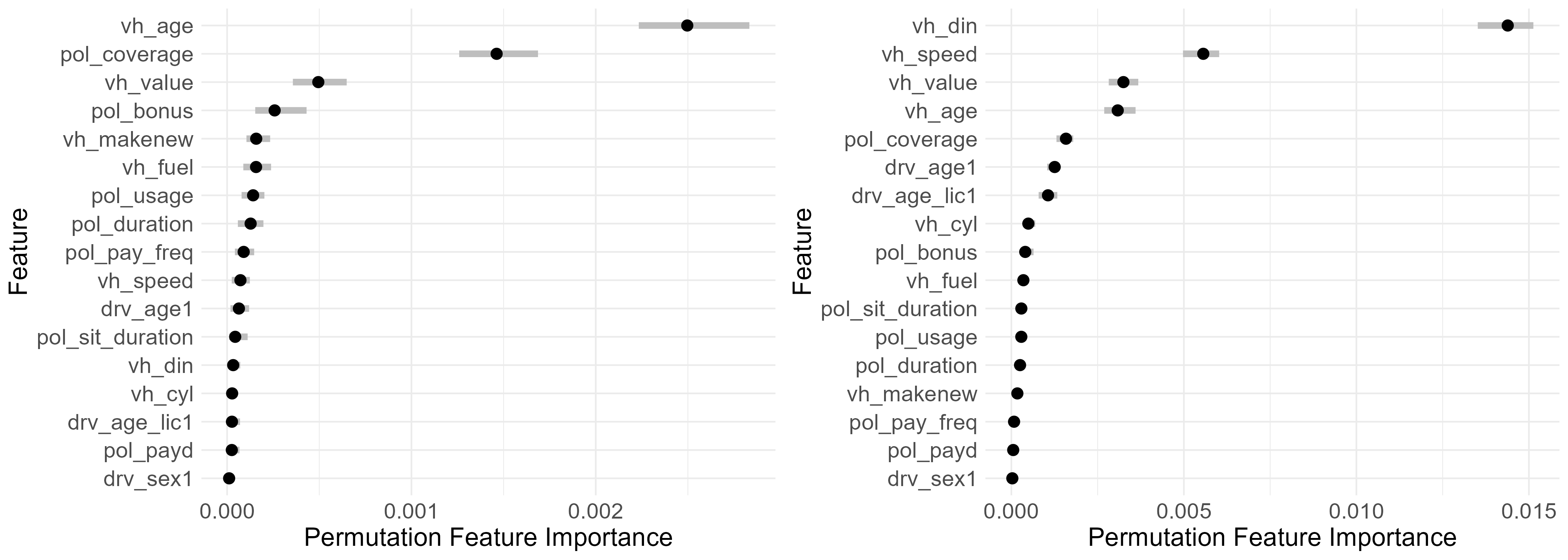}
    \caption{Permutation Feature Importance for $f(\textbf{x})$ before Attack (Left) and $a(\textbf{x})$ after Attack (Right), Insurance Data. The grey lines represent the range from the 10th to the 90th percentile of PFI estimates across fifty runs, with median values indicated by black dots.}
    \label{fig:pfi_pg17}
\end{figure}

\begin{figure}[h!]
    \centering
    \includegraphics[width=\textwidth]{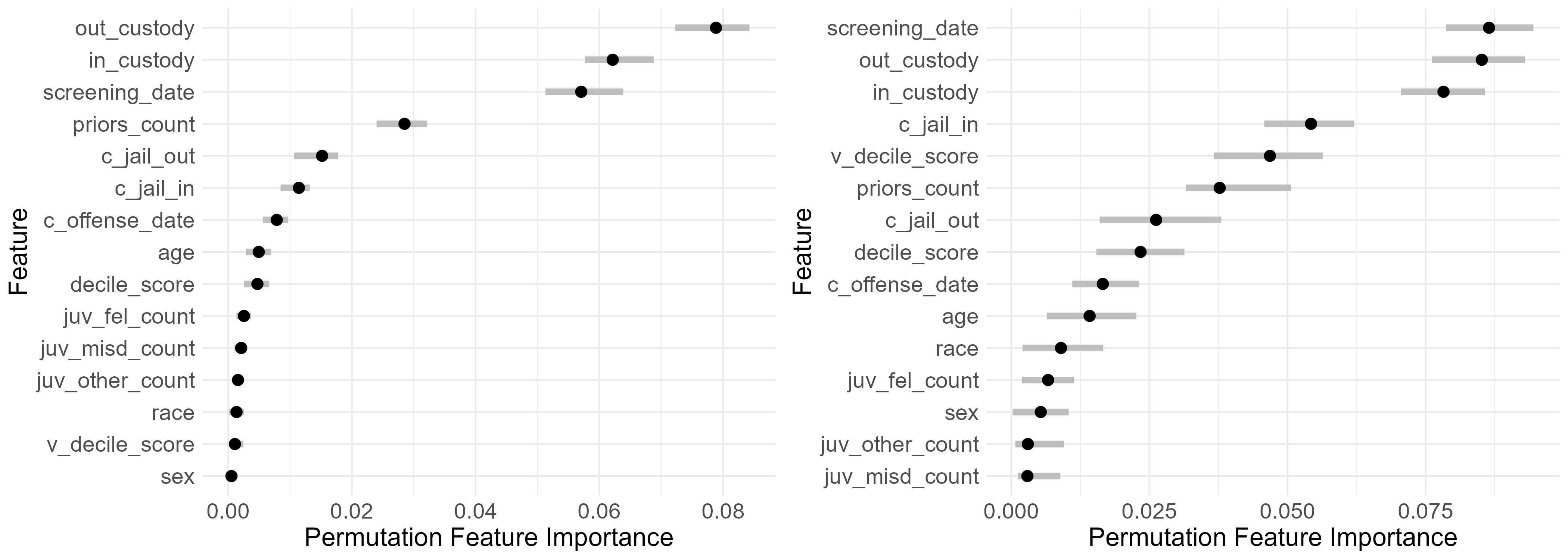}
    \caption{Permutation Feature Importance for $f(\textbf{x})$ before Attack (Left) and $a(\textbf{x})$ after Attack (Right), COMPAS Data. The grey lines represent the range from the 10th to the 90th percentile of PFI estimates across fifty runs, with median values indicated by black dots.}
    \label{fig:pfi_compas}
\end{figure}

In this appendix, we present the permutation feature importance of our fooling algorithm for the insurance and COMPAS datasets in Figures \cref{fig:pfi_pg17,fig:pfi_compas}.

\section{Alternatives to PD Plots}\label{sec:alternative}

In this appendix, we present several alternatives to PD plots aimed at addressing their limitations.

\textbf{Individual Conditional Expectation (ICE) plots}: ICE plots \citep{goldstein2015peeking} offer a more detailed view by breaking down the averages of PD plots into individual observations. Each ICE curve represents the conditional relationship between a range of $X_j$ values (where $|S| = 1$) and $\hat{f}$ while keeping other features $X_C$ fixed. Mathematically, it is defined as:
\begin{equation}
    \mathrm{ICE}^{(i)}_j (x) = \hat{f}(x^{(i)}_j = x, \textbf{x}^{(i)}_C)
\end{equation}
It should be noted that a PD curve ($\mathrm{PD}_j (x)$) is essentially the average of $n$ ICE curves. ICE plots are useful for identifying heterogeneity in relationships, demonstrating how different subgroups or individual data points respond to changes in the predictor variable. Moreover, similar ICE curves can be grouped to avoid ICE plots overplotting and offer a version of regional PD plots, thereby facilitating the identification of regions with less confounded feature effects \citep{britton2019vine, zhang2021interpreting, herbinger2022repid}.


\textbf{Accumulated Local Effect (ALE) plots}: ALE plots \citep{apley2020visualizing} average the local changes in the predictions and accumulate them across a grid of intervals. Similar to M-plots, ALE plots use conditional distributions $\mathbb{P}(X_C|X_S = x_S)$ rather than marginal distributions $\mathbb{P}(X_C)$ to mitigate the extrapolation issues present in PD plots. However, ALE plots can avoid the omitted nuisance variable bias, a limitation affecting the usefulness of M-plots. It is important to note that the interpretations of ALE plots are only locally valid within each interval.

In addition, compared to the recovery properties of PD plots, ALE plots adhere to additive $\hat{f}$ regardless of the correlation between $X_S$ and $X_C$, and they adhere to multiplicative $\hat{f}$ when $X_S$ and $X_C$ are independent \citep{apley2020visualizing}.

\textbf{Functional ANOVA Decomposition}: The concept of functional ANOVA decomposition as introduced by \citep{hooker2007generalized} is another avenue associated with PD plots. The PD function's counterparts can be derived by estimating $f_j(x_j)$ and $f_{-j}(\textbf{x}_{-j})$ to minimize the following expression:
\begin{align}
    \int (f(\textbf{x}) - f_j(x_j) - f_{-j}(\textbf{x}_{-j}))^2 p(\textbf{x}) d\textbf{x}
\end{align}
Here, $f_j(x_j)$ represents the individual effect for feature $X_j$, $f_{-j}(\textbf{x}_{-j})$ denotes an unknown function for all features except feature $X_j$, and $p(\textbf{x})$ represents the feature distribution. See also \citet{hiabu2023unifying}. It is worth noting that accurate estimation can be challenging, especially in high-dimensional settings.

\textbf{Confidence Intervals of PD Plots}: Alternatively, we can derive the confidence intervals of PD plots on bootstrap samples (for example, see \citet{cafri2016understanding}). \citet{molnar2021relating} suggested computing the confidence bands for PD plots by refitting models multiple times, thereby accounting for model variance.

\bibliographystyle{apacite}
\bibliography{reference, reference_1}

\end{document}